\documentclass[11pt]{elsarticle}

\makeatletter
\def\ps@pprintTitle{%
 \let\@oddfoot\@empty
 \let\@evenfoot\@empty
}
\makeatother

\usepackage[utf8]{inputenc}
\usepackage[a4paper]{geometry}
\usepackage{pdflscape} 
\usepackage{graphicx}
\usepackage{xcolor}
\usepackage{amssymb}
\usepackage{pifont}
\usepackage{enumitem}
\usepackage{threeparttable} 
\usepackage{longtable}
\usepackage{afterpage}
\usepackage{booktabs}
\usepackage{rotating}
\usepackage[linesnumbered, ruled, vlined]{algorithm2e} 
\usepackage{colortbl}
\usepackage{amsmath}
\usepackage{subcaption}
\usepackage{tabularx} 
\usepackage{multirow}

\usepackage{setspace}
\usepackage{natbib}
\usepackage{float}
\usepackage{array}

\usepackage[bookmarks=false]{hyperref}
\usepackage{cleveref}
\usepackage{etoolbox} 

\SetAlgoNlRelativeSize{-1} 
\SetAlCapFnt{\small} 
\SetAlCapNameFnt{\small} 
\DontPrintSemicolon

\newcommand{\eqnref}[1]{eq.~(\ref{#1})}
\newcommand{\Eqnref}[1]{Eq.~(\ref{#1})}

\bibpunct[, ]{(}{)}{,}{a}{}{,}

\newcolumntype{C}{>{\centering\arraybackslash}X}

\setlist[itemize]{itemsep=0pt, parsep=0pt, topsep=0pt, partopsep=0pt}
\usepackage{lineno} 

\AtEndEnvironment{equation}{\vspace{-0.25\baselineskip}}

\setpagewiselinenumbers
\modulolinenumbers[1]

\begin{document}
\onehalfspacing 
\begin{frontmatter}

\title{Integrated trucks assignment and scheduling problem with mixed service mode docks:  A Q-learning based adaptive large neighborhood search algorithm} 


\author[label1]{Yueyi Li}
\author[label2]{Mehrdad Mohammadi}
\author[label1]{\texorpdfstring{Xiaodong Zhang\corref{cor1}}{Xiaodong Zhang}}
\author[label1]{Yunxing Lan}
\author[label2]{Willem van Jaarsveld}

\affiliation[label1]{organization={Beijing Jiaotong University},
            addressline={ShangYuanCun NO.3}, 
            city={Beijing},
            postcode={100044},            
            country={China}}
            
\affiliation[label2]{organization={Eindhoven University of Technology},
            addressline={PO Box 513}, 
            city={Eindhoven},
            postcode={5600MB},             
            country={Netherlands}}
\cortext[cor1]{Corresponding author:Xiaodong Zhang (zhangxd@bjtu.edu.cn)}

\begin{abstract}
Mixed service mode docks enhance efficiency by flexibly handling both loading and unloading trucks in warehouses. However, existing research often predetermines the number and location of these docks prior to planning truck assignment and sequencing. This paper proposes a new model integrating dock mode decision, truck assignment, and scheduling, thus enabling adaptive dock mode arrangements. Specifically, we introduce a Q-learning-based adaptive large neighborhood search (Q-ALNS) algorithm to address the integrated problem. The algorithm adjusts dock modes via perturbation operators, while truck assignment and scheduling are solved using destroy and repair local search operators. Q-learning adaptively selects these operators based on their performance history and future gains, employing the epsilon-greedy strategy.
Extensive experimental results and statistical analysis indicate that the Q-ALNS benefits from efficient operator combinations and its adaptive mechanism, consistently outperforming benchmark algorithms in terms of optimality gap and Pareto front discovery. In comparison to the predetermined service mode, our adaptive strategy results in lower average tardiness and makespan, highlighting its superior adaptability to varying demands.

\end{abstract}







\end{frontmatter}



\section{Introduction}
\label{sec1}

Mixed Service Mode (MSM) is a flexible warehouse loading and unloading operation mode where a single dock can handle both inbound and outbound trucks. This is in contrast to the xclusive service mode, commonly seen in real-world terminals, where one side is designated for loading and the other for unloading. \citet{boysen_cross_2010} introduced this term in their review, and several scholars \citep{stephan_vis-a-vis_2011,bodnar_scheduling_2017,rijal_integrated_2019} have conducted related research on this mode. Their results indicate that MSM docks can effectively reduce waiting time and operation costs. While most current studies focus on the cross-dock scenario, this study is driven by the operational needs of real-world unmanned warehouses and highlights the potential for broader applications in various warehouse operations. In fact, the MSM is increasingly gaining industry attention\citep{ladier_cross-docking_2016}. On the one hand, it addresses the need for intensive dock utilization; on the other hand, concerns about potential operational confusion can be alleviated by intelligent warehousing systems and equipment, which makes the mode more feasible.

When employing the MSM in warehouse operations, besides deciding \textit{where and when} to schedule inbound and outbound trucks, another crucial decision is \textit{how} to arrange the dock modes. In current research, the arrangement of MSM docks is often predetermined. \citet{bodnar_scheduling_2017} pre-set the proportion of MSM docks among the total number of docks, and \citet{rijal_integrated_2019} pre-set the proportion and location of MSM docks. This study proposes a more flexible approach to the MSM, where the decision of each dock's mode is part of the decision-making process. This means adapting the dock mode based on the structure of inbound and outbound trucks. Hereafter, the problem is referred to as the Integrated Trucks Assignment and Scheduling Problem with Dock Mode Decision (TASP-DMD).

The TASP-DMD problem is a typical NP-hard combinatorial optimization problem for which meta-heuristic algorithms can provide promising solutions. Literature has effectively solved related problems using the Adaptive Large Neighborhood Search (ALNS) meta-heuristic algorithm. Considering the convenient framework and proven success of ALNS, this paper also adopts this algorithm. In ALNS mechanisms, significant potential for improvement lies particularly in the Operator Filtering (OF) and Adaptive Operator Selection (AOS)  \citep{gendreau_large_2019}. The OF process involves conducting comparative experiments to identify the most effective operators, thereby enhancing the overall performance. In ALNS, designing proper operators for specific problems is crucial \citep{windras_mara_survey_2022}. While many operators have been proposed, there is little analysis of the appropriate number or effectiveness of these operators. \citet{voigt_review_2024} tested and ranked operators in vehicle routing problems (VRP), showing that performance differences exist. Pre-set operators may perform well on similar instances but poorly on others with different characteristics. Thus, a thorough filtering of potential operators before their employment is necessary.
 The AOS process includes evaluating operator performance and selecting suitable ones for further exploration\citep{ropke_adaptive_2006,pisinger_general_2007}. This adaptiveness is achieved by continuously monitoring operator performance and adjusting the selection accordingly \citep{windras_mara_survey_2022}, ensuring the search remains efficient and adaptive. Classical AOS mechanism follows the roulette wheel principle as \citet{ropke_adaptive_2006}, while a few studied a stochastic universal sampling strategy \citep{chowdhury_modified_2019}. However, according to \citet{turkes_meta-analysis_2021}, the added value of simple methods for AOS is not significant, indicating that simple adaptive mechanisms might not be intelligent enough to focus on successful operators. Therefore, machine learning algorithms, which can rapidly capture environmental states and make decisions, have significant application potential in this area \citep{dorronsoro_graph_2023, aboutabit_new_2023}. This paper combines Q-learning, as a reinforcement learning algorithm, with ALNS, attempting to leverage the model-free and learning-based advantage of Q-learning to improve AOS performance.

Motivated by research gaps and practice value, this paper seeks to address several research questions:  1) How can the TASP-DMD problem be modeled and solved effectively? 2) Will the decision to incorporate dock mode outperform the predetermined dock mode, and to what extent? 3) How to identify better-performing operator combinations within ALNS? and 4) Does the integration of Q-learning enhance the performance of ALNS, and to what extent?

For the existing research, the contributions of this paper are:
\begin{itemize}
    \item We introduce new decision variables and constraints related to dock modes, integrating truck assignment, scheduling, and dock mode decisions. Comparative analysis with predetermined service modes demonstrates the model's adaptability to various demands and optimized dock utilization.
    \item Filtering better-performing operators for ALNS. We adopt a pairwise comparison method to identify better-performing operator combinations from a wide range of local search and perturbation operators. This OF process significantly improves the algorithm's performance in solving the TASP-DMD problem by focusing on the most effective operators. 
    \item Q-learning is embedded into the original ALNS framework to enhance the AOS process. By guiding decisions based on both historical operator performance and the current search state, the Q-Learning-based AOS mechanism makes more efficient choices. This results in superior performance when solving the TASP-DMD problem compared to benchmark algorithms, while maintaining comparable computational overhead.    
\end{itemize}

The remainder of the paper is structured as follows. Section \ref{sec2} presents an overview of existing research related to MSM docks scheduling, ALNS, and Q-learning. We formulate the problem in Section \ref{sec3} and present our Q-ALNS in Section \ref{sec4}. Section \ref{sec5} designs a set of experiments, and the computational results are reported in Section \ref{sec6}. This analysis leads to conclusions in Section \ref{sec7}.

\section{Literature review}
\label{sec2}

\subsection{Truck assignment and scheduling for mixed service mode docks}
Dock operational decisions focus on efficiently allocating limited dock resources to inbound and outbound trucks over a specific time period. Given the typical imbalance between the number of trucks and available docks, precise timing and sequencing are crucial to minimize waiting times for trucks on-site. From the comprehensive classification by \citet{buijs_synchronization_2014} and \citet{boysen_cross_2010}, it is evident that the assignment and scheduling of trucks are critical operational decisions for docks. Previous research often equated dock scheduling with truck assignment and scheduling, as the number and configuration of docks are typically predetermined at the tactical level. 
However, with MSM docks, this equivalence no longer holds, as decisions on dock modes are required.

A comparative analysis of the most relevant literature on truck assignment and scheduling with MSM docks, as summarized in Table \ref{tab:literature}, highlights several key points. 
Most research focused on cross-docking, aiming for seamless transfers between inbound and outbound trucks with minimal storage buffers \citep{ladier_cross-docking_2016}. However, storage buffers are necessary for many other warehouse scenarios like distribution centers (DCs), and dock scheduling significantly influences internal resource management \citep{wolff_internal_2021}. Only a few studies have considered internal resource constraints \citet{wolff_internal_2021,hermel_solution_2016}. Research on the impact of MSM dock scheduling on internal operational processes is insufficient.

Existing studies on MSM docks \citep{stephan_vis-a-vis_2011,shakeri_robust_2012,berghman_practical_2015,hermel_solution_2016} have not considered dock mode as a decision variable, often treating it as predetermined. 
Some studies, such as \citet{bodnar_scheduling_2017} and \citet{rijal_integrated_2019}, have recognized the benefits of MSM docks but treated the proportion of MSM docks as a fixed input. This approach limits the full flexibility of dock mode decisions and may lead to sub-optimal solutions.

Most related studies tend to employ heuristic-based algorithms. Recent research has begun using machine learning techniques to improve heuristic or exact algorithms. \citet{neamatian_monemi_dock_2023} used support vector machines to identify effective objective functions in different Benders decomposition stages, demonstrating the framework's effectiveness. Combining RL and other machine learning techniques with heuristic-based algorithms requires further research.
We aim to bridge these gaps by combining Q-learning, a model-free RL, with ALNS to improve solution efficiency and quality, specifically addressing TASP-DMD problems for DCs with MSM docks.

\begin{table}
    \centering
    \caption{Summary of the most related literature}
    \label{tab:literature}
    \scriptsize
    \setlength{\tabcolsep}{2pt} 
    \begin{threeparttable}
    \begin{tabular*}{\textwidth}{@{\extracolsep{\fill}} cclccccc @{}}
    \toprule
    \multirow{2}{*}{Publications} & \multicolumn{3}{c}{Problem characteristic} & \multirow{2}{*}{Objective functions} & \multirow{2}{*}{Mode Variable} & \multirow{2}{*}{Algorithms} \\
    \cmidrule(lr){2-4}
                                  & Scenarios & MSM & Handling &  & & \\
    \cmidrule(r){1-1} \cmidrule(r){5-7} 
    \citet{shakeri_robust_2012}            & C          & \multicolumn{1}{c}{Yes} & \multicolumn{1}{c}{Yes}    & makespan                  & No                            & two-phase HA                        \\
    \citet{berghman_practical_2015}            & DC          & \multicolumn{1}{c}{Yes} & \multicolumn{1}{c}{No}     & Tardiness, makespan                  & No                            & HA                        \\
    \citet{hermel_solution_2016}            & C          & \multicolumn{1}{c}{Yes} & \multicolumn{1}{c}{Yes}      & Transshipment, makespan                  & No                            & Hierarchal HA                        \\
    \citet{bodnar_scheduling_2017}            & C          & \multicolumn{1}{c}{Yes} & \multicolumn{1}{c}{No}     & Transshipment, tardiness                  & No                            & ALNS                        \\
    \citet{rijal_integrated_2019}             & C          & \multicolumn{1}{c}{Yes} & \multicolumn{1}{c}{No} &  Transshipment, tardiness, storage & No                            & ALNS                        \\
    \citet{wolff_internal_2021}            & C          &\multicolumn{1}{c}{No} & \multicolumn{1}{c}{Yes} & Internal resource requirements & No        & CG + HA          \\
    \citet{neamatian_monemi_machine_2023}            & C          & \multicolumn{1}{c}{No} & \multicolumn{1}{c}{No}  &  Transshipment & No                            & SVM+BD          \\  
    
    This paper                    & DC & \multicolumn{1}{c}{Yes} & \multicolumn{1}{c}{Yes} &  Tardiness, makespan, handling    & Yes                           & Q-learning + ALNS           \\ 
    \bottomrule
    \end{tabular*}
    \begin{tablenotes}
    \footnotesize
    \item \noindent MSM: mixed service mode dock; C: cross-docking; HA: heuristic algorithm; DC: distribution center; CG: column generation; ALNS: adaptive large neighborhood search; SVM: support vector machine; BD: benders decomposition.
    \end{tablenotes}
    \end{threeparttable}
\end{table}

\subsection{Original ALNS}

ALNS is a meta-heuristic algorithm derived from the Large Neighborhood Search (LNS) introduced by \citet{goos_using_1998}. ALNS enhances LNS by exploring large neighborhoods through multiple destroy and repair local search operators and an adaptive selection mechanism, enabling the convenient design of neighborhood structures based on decision content \citep{ropke_adaptive_2006}. A key element of ALNS is the careful design of appropriate operators, as well as the integration of an adaptive layer to dynamically select these operators during the search process \citep{windras_mara_survey_2022}.

Rigorous analysis and test of operators (among multiple operators) are crucial to the success of ALNS \citep{karimi-mamaghan_machine_2022,voigt_review_2024}. The flexibility of the ALNS framework often leads researchers to include a large number of operators, increasing the chances of finding better solutions. However, an inappropriate set of operators can diminish algorithm efficiency. Scholars \citep{windras_mara_survey_2022,turkes_meta-analysis_2021} emphasized the importance of thorough operator analysis. In a related study, \citet{voigt_review_2024} reviewed and ranked operators for the VRP problem and its variants, confirming variations in operator selection frequency. This paper underscores the necessity of evaluating and filtering operators tailored to specific problems. In the context of TASP-DMD problems, the abundance of operators requires careful filtering and selection based on their expected effectiveness. 

In ALNS, the adaptive operator selection (AOS) mechanism typically selects operators based on their weights, which are often updated using a score-based method. In this method, operators are assigned scores based on their recent performance, which are then used to adjust their weights and influence their probability of being selected in the future. For example, \citet{bodnar_scheduling_2017} assigned three types of scores to operators based on outcomes such as yielding a solution better than the global best, yielding a solution better than the current best, or no improvement. These scores are then used to update the weights of each operator accordingly. Similarly, \citet{rijal_integrated_2019} employed the same AOS mechanism for their study to handle dock allocation alongside truck scheduling. 
 However, the score-based method often overlooked differences in operator performance, as it simply classified performance levels and was often user-defined \citep{dorronsoro_graph_2023}. Studies like \citet{turkes_meta-analysis_2021} have pointed out that the simple AOS mechanisms used in most ALNS offer limited improvement to the algorithm’s performance. Developing effective AOS mechanisms remains an open challenge. Learning-based methods like ML techniques have emerged as promising alternatives \citep{karimi-mamaghan_machine_2022}.  \citet{dorronsoro_graph_2023} introduced a deep RL-based operator selection mechanism for the capacitated VRP, demonstrating that this learning-based mechanism outperforms both random and score-based methods. Compared to score-based method, learning-based methods can capture richer information related to operator selection, leading to better overall performance.

\subsection{Integration of Q-learning into ALNS}

Q-learning is a widely used RL algorithm that enables agents to learn optimal policies through interactions with the environment. Its model-free nature makes it suitable for complex and dynamic settings where constructing an accurate model is challenging or infeasible. 
When integrated into ALNS, Q-learning can extract valuable insights from the search process and guide subsequent searches, leading to more adaptive and efficient outcomes. In Q-learning-based AOS, Q-value updates replace the traditional weight updates, where operators were previously assigned scores, and now they are assigned rewards \citep{dorronsoro_graph_2023, aboutabit_new_2023}. Therefore, careful design of the reward scheme is crucial, as it determines the feedback each operator receives and directly influences the learning process. \citet{zhang_dynamic_2022} used deep Q-learning to handle uncertainties in synchronized transportation planning, showing that RL-based operators outperform randomly implemented ones. \citet{aboutabit_new_2023} incorporated Q-learning into the AOS mechanism of ALNS. Using the epsilon-greedy strategy, new operators are selected based on Q-value after reaching a local minimum. Computational results indicated that a Q-learning-based AOS mechanism enhances both the efficiency and average performance of the original ALNS.  However, both studies focused on the VRP, and neither tested the effectiveness of reward mechanisms. 
In a related field, \citet{karimi-mamaghan_learning_2023} combined Q-learning with the iteration greedy algorithm. They proposed a value-based AOS mechanism, which differs from the score-based method by directly updating Q-values based on the amount of improvement an operator achieves. Their approach demonstrates improvements in both solution quality and convergence rate, serving as an important inspiration for this work.

\section{Problem description and formulation}
\label{sec3}
\subsection{Problem description}

\begin{figure}[ht] 
\centering
\includegraphics[scale=0.5]{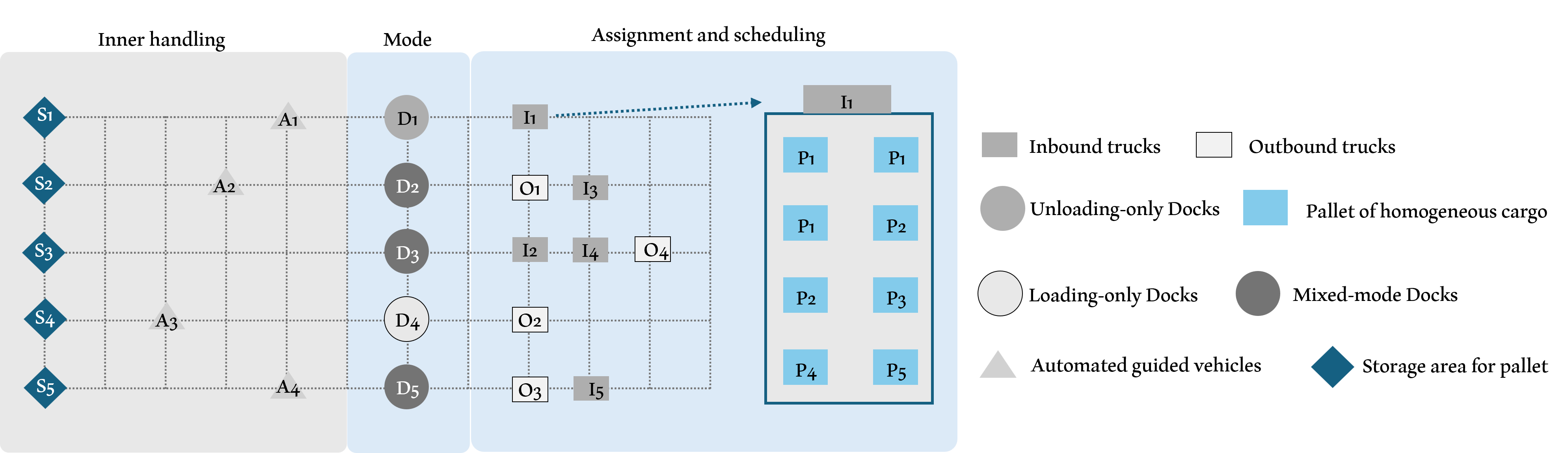} 
\caption{Truck assignment and scheduling problem with mixed-mode dock} 
\label{fig:problem define} 
\end{figure}

We consider an unmanned distribution center (UDC) that needs to allocate a set of trucks, \(T\), including inbound \(I\) and outbound \(O\) trucks. Inbound trucks \(i \in I\) can be assigned to either the unloading-only or mixed-mode docks. Outbound trucks \(j \in O\) can be allocated to either the loading-only or mixed-mode docks. As depicted in Figure~\ref{fig:problem define}, the UDC is equipped with several dock doors \(D\). Each dock \(k \in D\) can operate in one of the three modes: loading-only (\(k_i\)), unloading-only (\(k_j\)), or mixed-mode (\(k_m\)).

Trucks' arrivals are scheduled in advance and have fixed arrival and expected departure times, denoted by time windows \((r_i, d_i)\). The lower bound of the time window cannot be moved earlier, but the upper bound can be extended, as departure times may extend beyond scheduled windows due to resource constraints. All cargo is palletized on standard-sized pallets, with each pallet containing homogeneous cargo, denoted as \(P\). The types and quantities of pallets \(p \in P\) for inbound trucks are known, as are the types and quantities of pallets that need to be loaded onto outbound trucks. The UDC has a storage area \(S\) where the storage location \(s \in S\) of each pallet \(u_{ps}\) is known. Based on the design capacity of the UDC, we assume that storage space constraints will not be exceeded.
Inbound trucks, once assigned to a dock, need to unload pallets, which are then handled in the corresponding storage area by Automated Guided Vehicles (AGVs). Conversely, for outbound trucks, AGVs handle pallets from the storage area to the assigned dock for loading onto the trucks. Most research does not consider the impact of internal cargo transport, focusing more on the completion time of tasks at the truck end. However, in resource-limited situations, the efficiency of internal transport is a crucial factor for overall warehouse efficiency \citep{wolff_internal_2021}. Therefore, including AGV transport distance as an objective quantifies the impact of each decision on the efficiency of the entire inbound and outbound process and controls the scheduling process with a focus on both transportation and warehouse system needs.
Additionally, due to the switch between loading and unloading tasks, MSM docks require a reaction time \(\tau\) before a task begins.

The scheduling decisions encompass the mode of the dock, truck-to-dock assignment, and truck scheduling, collectively termed the Integrated Truck Assignment and Scheduling Problem with Dock Mode Decision (TASP-DMD). The optimization objectives include minimizing truck delay, the maximum completion time, and the weighted distance of cargo handling. The truck delay accounts for any extensions beyond the scheduled departure time \(d_i\), while the weighted distance of cargo handling considers the distance between the dock and the storage area. 

Several assumptions have been made in this paper based on practical cases and experience: 1) Each dock is designed to be equipped with sufficient forklifts, and the travel paths of the forklifts are fixed, simplifying loading and unloading time to the time per pallet, denoted as \(t_e\); 2) Trucks can only begin operations after their scheduled arrival time; 3) A truck can leave the docking position only after completing its loading or unloading tasks; 4) No ad-hoc truck loading or unloading tasks are included; and 5) Compatibility between docking positions and truck types is not considered.

\subsection{Notations}
Before proposing the mathematical model of the problem, parameters and decision variables are listed in Table~\ref{tab:notation}.

\subsection{Truck assignment and scheduling model with dock mode decision}
\label{model}
 
This section constructs a multi-objective mixed-integer programming model that simultaneously optimizes dock mode, truck assignment, and truck scheduling. The optimization objectives are to minimize dock operation delay time (referred to as tardiness), the maximum completion time (referred to as makespan), and the distance of cargo handling (referred to as distance).

\small
\begin{longtable}{cp{12cm}}
\caption{Notation Definitions for TASP-DMD Problem} \label{tab:notation} \\
\toprule
Symbol & Notation Definition\\
\midrule
\endfirsthead

\multicolumn{2}{c}{{\bfseries \tablename\ \thetable{} -- continued from previous page}} \\
\toprule
Symbol & Notation Definition\\
\midrule
\endhead

\midrule \multicolumn{2}{r}{{Continued on next page}} \\
\endfoot

\bottomrule
\endlastfoot

\textbf{Parameters} & \\
\(r_i\) & Scheduled arrival time of truck \(i\), \(\forall i \in T\). \\
\(d_i\) & Expected departure time of truck \(i\), \(\forall i \in T\). \\
\(t_e\) & Loading and unloading time per unit of cargo. \\
\(\tau\) & Additional waiting time required before each truck starts loading and unloading operations at mixed-mode docks. \\
\(a_i\) & Start time of operations for truck \(i\), \(\forall i \in T\). \\
\(e_i\) & End time of operations for truck \(i\), \(\forall i \in T\). \\
\(h_{ip}\) & Number of pallets of cargo type \(p\) to be unloaded or loaded on truck \(i\), \(\forall p \in P\), \(\forall i \in T\). \\
\(A_i\) & Total quantity of cargo to be unloaded or loaded on truck \(i\), \(A_i = \sum_{p \in P} h_{ip}\). \\
\(v\) & Travel speed of the handling equipment provided at each dock. \\
\(u_{ps}\) & is \(1\) when cargo of type \(p\) are stored in area \(s\), and is \(0\) otherwise. \\
\(N_D\) & Number of docks. \\
\(H_i\) & Total time for truck \(i\)'s loading or unloading tasks \\
\(H_{max}\) & Number of docks. \\
\addlinespace
\textbf{Decision Variables} & \\
\(x_{ik}\) & is \(1\) if truck \(i\) designated for unloading is assigned to operate at dock \(k\), and is \(0\) otherwise \\
\(y_{jk}\) & is \(1\) if truck \(j\) designated for loading is assigned to operate at dock \(k\), and is \(0\) otherwise \\
\(\lambda_k\) & is \(1\) if dock \(k\) is exclusively for unloading, and is \(0\) otherwise \\
\(\rho_k\) & is \(1\) if dock \(k\) is exclusively for loading, and is \(0\) otherwise \\
\(\mu_k\) & is \(1\) if dock \(k\) is for mixed-mode, and is \(0\) otherwise \\
\(z_{ijk}\) & is \(1\) if truck \(i\) and truck \(j\) are assigned to dock \(k\), and \(i\) is the preceding truck in the sequence before \(j\), and is \(0\) otherwise \\
\addlinespace
\textbf{Auxiliary Variables} & \\
\(f_{ik}\) &  is \(1\) if truck \(i\) is first in the sequence at dock \(k\), and is \(0\) otherwise \\
\(l_{ik}\) & is \(1\) if truck \(i\) is last in the sequence at dock \(k\), and is \(0\) otherwise \\
\end{longtable}


The tardiness refers to the sum of the delays of all trucks' operations compared to their scheduled departure times. The expression for the first objective function is shown in \eqnref{eq:obj1}. Here, \(\delta_i\) represents the delay time for truck \(i\), defined as \(\delta_i = \max \{ 0, e_i - d_i \}\), where \(e_i\) is the time when truck \(i\) finishes its operations.

\small
\begin{equation}
\ f_1 = \min \sum_{i \in I \cup O} \delta_i
\label{eq:obj1}
\end{equation}

The makespan refers to the latest time by which all loading and unloading tasks are completed. The expression for the second objective function is presented in \eqnref{eq:obj2}.

\small
\begin{equation}
\ f_2 = \min \max \{ e_i \mid i \in I \cup O \}
\label{eq:obj2}
\end{equation}

Assigning trucks to different docking positions results in varying distances for internal handling. The third objective function aims to minimize the total distance for handling. The expression for this objective function is detailed in \eqnref{eq:obj3}. Detailing the scheduling of AGVs would significantly increase the complexity of the problem. Therefore, we adopt the Euclidean distance between dock positions and storage locations. The distance \(m_{ks}\) between dock position \(k\) and storage location \(s\) is given by \eqnref{eq:distance}, where \(\Delta x_{ks}\) and \(\Delta y_{ks}\) are the horizontal and vertical distances between them, respectively. Detailed scheduling will be addressed in future research. 

\vspace{-10pt}
\begin{align}
& f_3 = \min \sum_{i \in I, j \in O} \sum_{k \in D} \sum_{s \in S} \sum_{p \in P} x_{ik} h_{ip} u_{ps} m_{ks} \label{eq:obj3} \\
& m_{ks} = \sqrt{(\Delta x_{ks})^2 + (\Delta y_{ks})^2} \label{eq:distance}
\end{align}
\vspace{-10pt}

The first category of constraints pertains to the allocation of dock positions. \Eqnref{eq:dock_service_type_constraint} ensures that each dock is designated for only one service type: unloading-only (\(\lambda_k\)), loading-only (\(\rho_k\)), or mixed-mode (\(\mu_k\)).
\Eqnref{eq:inbound_dock_assignment} and \eqnref{eq:outbound_dock_assignment} specify that each truck is assigned to only one dock. \Eqnref{eq:inbound_dock_capability} ensures that unloading trucks are assigned only to docks that are capable of handling unloading, specifically exclusive unloading docks or MSM docks. \Eqnref{eq:outbound_dock_capability} applies a similar logic to outbound trucks.

\vspace{-10pt}
\begin{align}
&\lambda_k + \mu_k + \rho_k = 1, \forall k \in D \label{eq:dock_service_type_constraint} \\
&\sum_{k \in D} x_{ik} = 1, \forall i \in I \label{eq:inbound_dock_assignment} \\
&\sum_{k \in D} y_{jk} = 1, \forall j \in O \label{eq:outbound_dock_assignment} \\
&\sum_{k \in D} (\lambda_k x_{ik} + \mu_k x_{ik}) = 1, \forall i \in I \label{eq:inbound_dock_capability} \\
&\sum_{k \in D} (\rho_k y_{jk} + \mu_k y_{jk}) = 1, \forall j \in O \label{eq:outbound_dock_capability}
\end{align}

The second category of constraints deals with the timing of truck loading and unloading operations. \Eqnref{eq:timing_sequence} defines the sequential relationship in time for trucks serviced at each docking position, with \(H_i\) representing the time for inbound and outbound operations. The value is determined as in \eqnref{eq:mixed_mode_time}. Here, the first term accounts for the forklift's loading and unloading time, the second term for AGV handling time, and the third term for the preparation time required at mixed-mode docks. The sum includes all these times only when both \(\mu_k\) and \(x_{ik}\) are set to 1. \Eqnref{eq:max_completion_time} and \eqnref{eq:arrival_time_constraint} represent the time window constraints.
\begin{align}
&a_i + H_i \leq a_j + M(1 - z_{ijk}), \forall i, j \in I \cup O, k \in D \label{eq:timing_sequence} \\
&a_i + H_i \leq H_{\max}, \forall i \in I \cup O \label{eq:max_completion_time} \\
&a_i \geq r_i, \forall i \in I \cup O \label{eq:arrival_time_constraint} \\
&H_i = \frac{A_i}{t_e} + \sum_{g \in G} \sum_{s \in S} \sum_{k \in D} x_{ik} h_{ip} u_{ps} m_{ks} \cdot \frac{1}{v} + \sum_{k \in D} \mu_k x_{ik} \tau \label{eq:mixed_mode_time}
\end{align}

The third category of constraints pertains to the sequencing or ordering of trucks. \Eqnref{eq:sequencing_conflict} addresses the potential confusion in the assignment of truck sequencing variables, stipulating that if truck \(i\) precedes truck \(j\) at the same dock, then \(j\) cannot precede \(i\) at that dock. \Eqnref{eq:sequencing_dependency} indicates that \(z_{ijk}\) can only be 1 if trucks \(i\) and \(j\) are assigned to the same dock. \Eqnref{eq:mixed_mode_assignment} ensures that trucks \(i\) (unloading) and \(j\) (loading) are assigned to the same dock only when \(\mu_k = 1\), providing a linear representation of the constraint \(x_{ik} y_{jk} \leq \mu_k\). \Eqnref{eq:first_sequence} to \eqnref{eq:only_last_sequence} define the constraints for when a truck's loading or unloading sequence is either first or last at a dock, utilizing auxiliary variables \(f_{ik}\) and \(l_{ik}\), which are set to 1 when a truck is the first or last in sequence at dock \(k\), and 0 otherwise.

\vspace{-10pt}
\begin{align}
&z_{ijk} + z_{jik} \leq 1, \forall i, j \in I \cup O, k \in D \label{eq:sequencing_conflict} \\
&z_{ijk} \leq x_{ik} x_{jk}, \forall i, j \in I \cup O, k \in D \label{eq:sequencing_dependency} \\
&x_{ik} + y_{jk} - 1 \leq \mu_k, \forall i \in I, j \in O, k \in D \label{eq:mixed_mode_assignment} \\
&f_{ik} + \sum_{j \in I \cup O \setminus i} z_{jik} = x_{ik}, \forall i \in I \cup O, k \in D \label{eq:first_sequence} \\
&l_{ik} + \sum_{j \in I \cup O \setminus i} z_{ijk} = x_{ik}, \forall i \in I \cup O, k \in D \label{eq:last_sequence} \\
&\sum_{i \in I \cup O} f_{ik} = 1, \forall k \in D \label{eq:only_first_sequence} \\
&\sum_{i \in I \cup O} l_{ik} = 1, \forall k \in D \label{eq:only_last_sequence}
\end{align}

The final category of constraints pertains to the types of variables.

\vspace{-10pt}
\begin{align}
&x_{ik}, y_{jk} \in \{0,1\}, \forall i \in I, j \in O, k \in D \label{eq:binary_decision_vars} \\
&\lambda_k, \mu_k, \rho_k \in \{0,1\}, \forall k \in D \label{eq:binary_dock_mode_vars} \\
&z_{ijk} \in \{0,1\}, \forall i, j \in I \cup O, k \in D \label{eq:binary_sequence_vars} \\
&f_{ik}, l_{ik} \in \{0,1\}, \forall i \in I \cup O, k \in D \label{eq:binary_auxiliary_vars} \\
&a_i \geq 0, \forall i \in I \cup O \label{eq:nonnegative_start_time}
\end{align}

The model, implemented in Python 3.8 and utilizing the Gurobi 11.0.1, was evaluated to demonstrate its practicality and accuracy. The test data includes six docks and twenty trucks, where eight trucks are for unloading and twelve are for loading. 
The model was tested on a computer with an Intel® Core(TM) i7-10210U processor, 1.6GHz, 4GB memory, running Windows 11, using the Visual Studio platform. The solution process took 1.23 seconds, yielding optimal values $f_1=0$, $f_2=105.06$, and $f_3=3818.5$ The results are presented in Table~\ref{table:Gurobi_results}.

\begin{table}[ht]
    \centering
    \scriptsize
    \caption{Optimal solution for a small-scale example}
    \label{table:Gurobi_results}
    \begin{tabular}{ccc|ccc}
        \hline
        Dock & Mode & Truck order & Dock & Mode & Truck order \\
        \hline
        $D_1$ & Unloading-only & 1 $\rightarrow$ 6 & $D_4$ & Loading-only & 13 $\rightarrow$ 15 $\rightarrow$ 17 $\rightarrow$ 12 $\rightarrow$ 10 \\
        $D_2$ & Unloading-only & 7 $\rightarrow$ 2 & $D_5$ & Loading-only & 16 $\rightarrow$ 9 $\rightarrow$ 11 $\rightarrow$ 20 \\
        $D_3$ & Mixed-mode & 3 $\rightarrow$ 5 $\rightarrow$ 8 $\rightarrow$ 4 & $D_6$ & Loading-only & 14 $\rightarrow$ 18 $\rightarrow$ 19 \\
        \hline
    \end{tabular}
\end{table}

\section{Proposed ALNS with a Q-learning-based AOS: Q-ALNS}
\label{sec4}

\begin{figure}[ht] 
\centering
\includegraphics[scale=0.4]{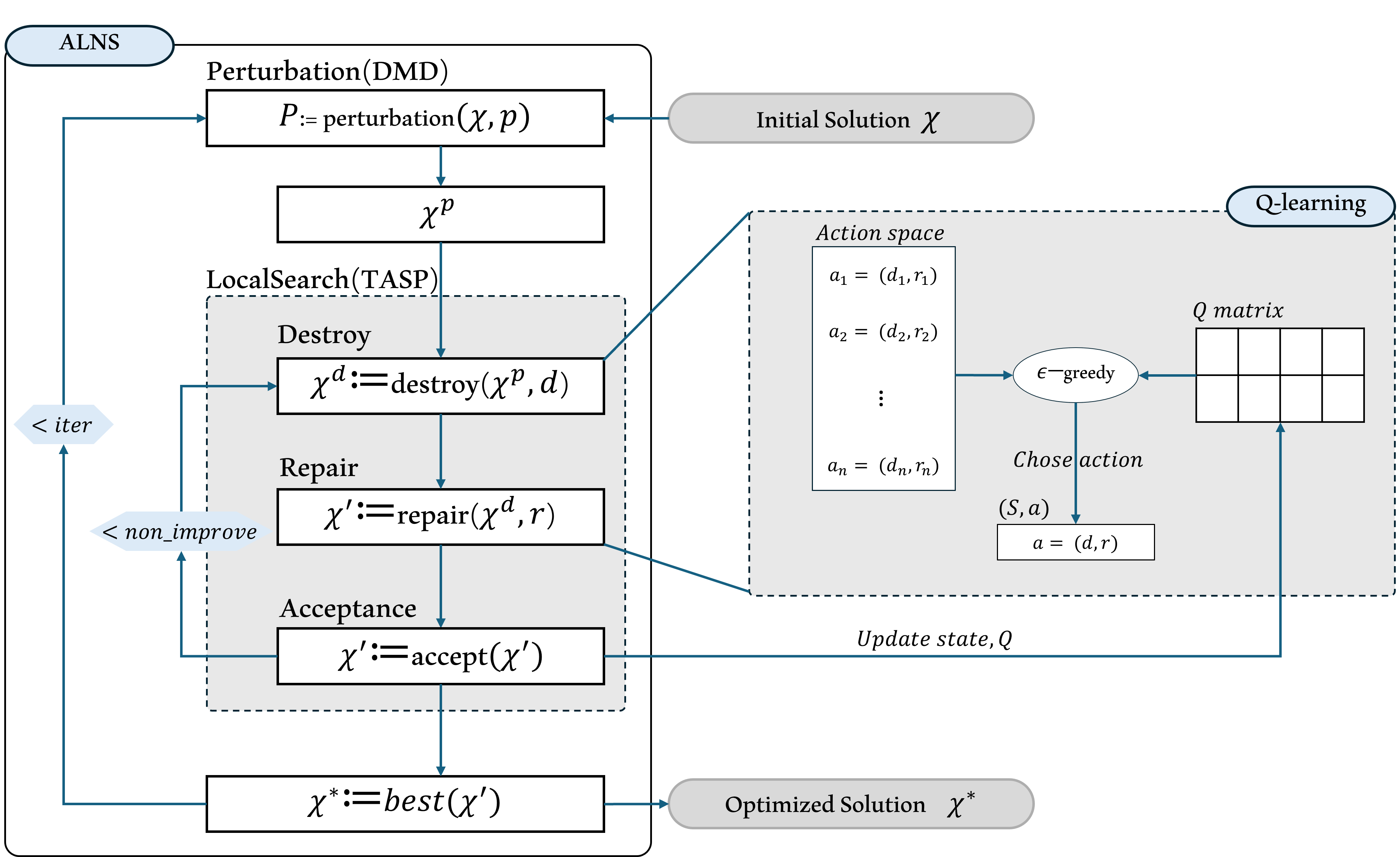} 
\caption{Flowchart of the Q-ANLS} 
\label{fig:outline} 
\end{figure}

Figure~\ref{fig:outline} illustrates the flowchart of the proposed Q-ALNS. Additionally, Algorithm~\ref{alg:qalns} provides the detailed Pseudo-code of the Q-ALNS.
The main improvements of the proposed Q-ALNS over the original ALNS are as follows: 

1) The main loop is extended into a two-layer loop. The outer loop employs perturbation operators to achieve diverse neighborhood searches for dock mode decisions. The inner loop uses destroy and repair local search operators for truck assignment and scheduling under the new dock mode. This approach allows for the simultaneous optimization of dock mode, truck assignment, and scheduling. 

2) The Q-learning algorithm is integrated into the AOS mechanism, which bases its decisions on the search state and history of performance. The gray areas in Figure~\ref{fig:outline} illustrate this mechanism. During the local search phase of the algorithm, each combination of destroy and repair operators forms an action. After applying an action, a corresponding reward is immediately obtained based on the improvement in the solution quality. Then, the Q-learning algorithm assigns a credit to this action and updates the Q-table. Unlike score-based methods, Q-learning also accounts for future gains, encouraging long-term benefits. The next action is selected using the $\epsilon$-greedy strategy. This strategy uses the Q-table to choose actions with the highest credit or explore new actions, preventing the algorithm from getting stuck in local optima.   

3) Since we are dealing with a multi-objective model, our algorithm updates solutions by maintaining and updating a Pareto set of solutions. A new solution is accepted under one of the following four conditions: when the new Pareto solution dominates the current global best solution, the global best solution is updated; when the new Pareto solution dominates the current local best solution, the local best solution is updated; when the new solution and the current global best solution are mutually non-dominating, the new solution is added to the Pareto set; and when the new solution does not meet any of the above criteria, it is accepted according to the Metropolis acceptance function \cite{karimi-mamaghan_learning_2023}.

\begin{algorithm}
    \caption{Pseudo-code for ALNS with Q-learning (Q-ALNS)}
    \label{alg:qalns}
    \footnotesize

    \KwIn{$p$ \tcp*{Perturbation operators}}
    \KwIn{$op$ \tcp*{Local search operators}}
    \KwIn{$A$ \tcp*{Action space}}
   
    \KwIn{$\epsilon \in \mathbb{R}$ \tcp*{Epsilon-greedy parameter}}
    \KwIn{$\beta \in \mathbb{R}$ \tcp*{Epsilon-decay}}
    \KwIn{$\alpha \in \mathbb{R}$ \tcp*{Learning rate}}
    \KwIn{$\gamma \in \mathbb{R}$ \tcp*{Discount factor}}
    \KwIn{$E \in \mathbb{N}$ \tcp*{Size of iteration}}
    \KwIn{$L\in \mathbb{N}$ \tcp*{Size of learning loop}} 
    \KwIn{$\eta \in \mathbb{R}$ \tcp*{Global improvement weight}}

    \KwIn{$t \in \mathbb{N}$ \tcp*{Increment non\_improve counter}}
    \KwOut{$\chi^*$ \tcp*{Optimized solution found}}
    \SetKwFunction{ALNS}{ALNS}
    \SetKwProg{Fn}{Function}{:}{}
    
    \Fn{\ALNS{$\epsilon, \beta, \alpha, \gamma, E, \eta, A, t $}}{
        $\chi \leftarrow \text{initialsolution()}$ \tcp*{Construct initial solution}
        $\chi^* \leftarrow \chi$ \tcp*{Initialize best solution}
        $Q \leftarrow [0]$ \tcp*{Initialize Q-table}
        $s \leftarrow 0$ \tcp*{Initialize state}
        $a \leftarrow \text{randomchoice}(A)$ \tcp*{Initialize action}
        
        \tcp{Outer loop}
        \For{$e \leftarrow 1$ \KwTo $E$} {
            $R_{\text{prev}} \leftarrow R_{\min}(\chi)$ \tcp*{Remember current local optimum before a local search episode}
            $R_{\text{prev}}^* \leftarrow R_{\min}(\chi^*)$ \tcp*{Remember the best solution found before a local search episode}
            $R \leftarrow R_{\text{prev}}$ \tcp*{Remember the best local optimum found during an episode}
            $R^* \leftarrow R_{\text{prev}}^*$ \tcp*{Remember the best solution found during an episode}
            \uIf{$e < L$}{
                $a \leftarrow \text{randomchoice}(A)$ \tcp*{Action is drawn randomly from $A$}
            }
            \Else{
                $a \leftarrow \text{Q-learning}(R_{\text{prev}}, R_{\text{prev}}^*, R, R^*, \epsilon, \beta, \alpha, \gamma, E, \eta, A, s, \text{op})$ \tcp*{Action based on Q-table}
            }
            \tcp{Inner loop}
            $\texttt{non\_improve} \leftarrow 0$ \tcp*{Initialize non\_improve counter}
            $T \leftarrow -\sum f_k(\text{sol}) / \log(0.5)$ \tcp*{Initialize temperature}
            \While{\texttt{non\_improve} $<$ \texttt{t}}{                   
                \tcp{destroy and repair phase}
                $\chi' \leftarrow N(\chi, \text{op})$ \tcp*{local search episode}
                \tcp {Update best solution found}
                \If{$\text{Pareto dominate}(\chi', \chi^*)$}{
                    $\chi^* \leftarrow \chi'$ \\
                    $R^* \leftarrow R_{\min}(\chi^*)$
                }
                \ElseIf{$\text{Pareto dominate}(\chi', \chi)$}{
                    $\chi \leftarrow \chi'$ \\
                    $R \leftarrow R_{\min}(\chi)$ \\                    
                }
                \ElseIf{$\text{mutually non-dominating}(\chi', \chi^*)$}{
                    \texttt{add $\chi'$ to Pareto set} \\                    
                }
                \ElseIf{$\text{accept}(\chi', \text{temperature criterion})$}{
                    $\chi \leftarrow \chi'$ \\
                    \texttt{non\_improve} $\leftarrow \texttt{non\_improve} + 1$
                }
                \Else{
                    \texttt{non\_improve} $\leftarrow \texttt{non\_improve} + 1$
                }
                update $T $ \tcp*{Update temperature}
            }
            
            update $Q, s$ \tcp*{Update q-table}
            $\chi \leftarrow P(\chi^{*}, p)$ \tcp*{Do perturbation}
        }
        \Return{$\chi^*$} \tcp*{Return the best solution}
    }
\end{algorithm}

\subsection{States and actions in the proposed Q-ALNS}
In the proposed Q-ALNS, Q-learning is utilized to select local search operators at each iteration, as illustrated in Algorithm~\ref{alg:qlearning}. The set of actions is correspondingly defined to determine which local search operator to apply, represented as \eqnref{eq:action}, where \( a \) is the number of operators. Each operator is associated with a set of destroy and repair operations. Upon selecting an action, the corresponding operations are performed on the current solution \( \chi \), leading to a new solution \( \chi' \).

\begin{algorithm}
    \caption{Pseudo-code for Q-learning Function}
    \label{alg:qlearning}
    \scriptsize
    
    \KwIn{$R_{\text{prev}}$, $R_{\text{prev}}^*$, $R$, $R^*$, $\epsilon$, $\beta$, $\alpha$, $\gamma$, $E$, $\eta$, $A$, $s$, $op$}
    \KwOut{$Q$, $s'$, $a'$ \tcp*{New state $s'$ and next action $a'$}}
    \SetKwFunction{QL}{QL}
    \SetKwProg{Fn}{Function}{:}{}
    \Fn{\text{Q-learning}($R_{\text{prev}}, R_{\text{prev}}^*, R, R^*, \epsilon, \beta, \alpha, \gamma, E, \eta, A, s, op$)}{
        \tcp{Calculate reward based on equations}
        $r \leftarrow \text{CalculateReward}(R_{\text{prev}}, R_{\text{prev}}^*, R, R^*, e, \eta)$ \;
        
        \tcp{Determine state based on improvement during the iteration}
        \If{$R^* < R_{\text{prev}}^*$}{
            $s' \leftarrow 1$ \;
        }
        \Else{
            $s' \leftarrow 0$ \;
        }
        
        \tcp{Update Q-table}
        
        $ Q(s, a) \leftarrow  Q(s, a) + \alpha \left( r + \gamma \max_{a'} Q(s', a') - Q(s, a) \right) $\;

        \tcp{Apply epsilon-decay strategy to move from exploration to exploitation gradually}
        $\epsilon \leftarrow \epsilon \cdot \beta$ \;
        
        \tcp{Choose an action using $\epsilon$-greedy strategy}
        \If{$\text{rand}() \geq \epsilon$}{
            $a' \leftarrow \text{argmax}_{a''} Q(s', a'')$ \;
        }
        \Else{
            $a' \leftarrow \text{randomchoice}(A)$ \;
        }
        
        \Return{$Q, s', a'$}
    }
\end{algorithm}

\small
\begin{equation}
 A = \{ 1, 2, \ldots, a \}
\label{eq:action}
\end{equation}

In Q-learning, a state represents the condition or situation of the environment at a particular time. It is a critical factor in deciding the action to be taken to achieve the optimal solution. The state is determined by evaluating whether the current solution shows any improvement, thus indicating whether the algorithm is trapped in a local optimum. Therefore, in the Q-learning mechanism of this study, the state is set as a binary set \( S = \{ 0, 1 \} \). If new state \( s' = 0 \), it implies that the local search in the latest iteration could not find a better solution. Conversely, if \( s' = 1 \), it means that a better solution was found after the latest iteration.
 Let \(\Delta\) denotes the improvement measure, the state transition function \(P(s' \mid s, a)\) based on this rule is then as follows:

\small
\begin{equation}
P(s' \mid s, a) = 
\begin{cases} 
1 & \text{if } \Delta > 0 \text{ and } s' = 1 \\
1 & \text{if } \Delta \leq 0 \text{ and } s' = 0 \\
0 & \text{otherwise}
\end{cases}
\label{eq:trans}
\end{equation}

In each iteration of the Q-ANLS, the performance of the local search operators is evaluated, and the reward is calculated accordingly. The Q-table is then updated based on this reward and the current $s$, as \eqnref{eq:updateQ}. Finally, using the $\epsilon $-greedy strategy, the next local search operator is selected based on the current state and the updated Q-table.

\small
\begin{equation}
Q(s,a) = Q(s,a) + \alpha \left( r + \gamma \max_{a'} Q(s',a') - Q(s,a) \right)
\label{eq:updateQ}
\end{equation}

\subsection{Reward function in the proposed Q-ALNS}
\label{sec4.2}

The reward in Q-learning represents the feedback received after taking an action, reflecting the benefit of that action in improving the current solution. In each iteration of the Q-ALNS, a local search operator is selected, and a local search is performed until no further improvement is achieved. Thus, the evaluation of the performance of a local search operator is conducted at the end of each iteration. 
The reward can be defined in various ways. In this work, we reference the definition proposed by \citet{karimi-mamaghan_learning_2023} to avoid the algorithm getting stuck in local optima and accelerate convergence. In their work, the reward is a function of global and local improvements of the solutions. Building on their methods, we incorporate the consideration of iteration count to assign higher values to greater improvements at later iterations, as illustrated in Algorithm \ref{alg:qlearning_reward}.

The improvement in the solution includes enhancements in both global and local optima, computed using \eqnref{eq:global} and \eqnref{eq:local}, respectively. These improvements are combined linearly
in \eqnref{eq:weighted}, and multiplied by the iteration count to calculate the reward \( r \) (see \eqnref{eq:reward}), where \(e\) is the iteration count and  \( C \) is a normalization constant. Due to the difference in magnitude between iteration counts and improvement values, the result is normalized by dividing by a constant. A penalty mechanism was also incorporated: if no improvement is obtained, the reward is adjusted to include the opportunity cost 
\( \text{OC} \), reflecting the maximum prior improvement of unselected local search operators, as described in \eqnref{eq:reward-punish}.  

\begin{align}
&\Delta \text{Global} = \max \left(\frac{R^* - R^*_{prev}}{R^*_{prev}}, 0\right) \label{eq:global} \\
&\Delta \text{Local} = \max \left(\frac{R - R_{prev}}{R_{prev}}, 0\right) \label{eq:local} \\
&\Delta \text{Improvement} = \Delta \text{Global} \times \eta + \Delta \text{Local} \times (1 - \eta) \label{eq:weighted} \\
&r_1 = \frac{\Delta \text{Improvement} \times e}{C} \label{eq:reward} \\
&r_2 = \frac{(\Delta \text{Improvement} - \text{OC}) \times e}{C} \label{eq:reward-punish}
\end{align}

\begin{algorithm}[ht]
    \caption{Q-learning Reward Calculation with Penalty Mechanism}
    \label{alg:qlearning_reward}
    \scriptsize
    \KwIn{$R^*$, $R^*_{prev}$, $R$, $R_{prev}$, $\eta$, $e$, $C$}
    \KwOut{$r$ \tcp*{Reward}}
    
    \SetKwFunction{CalculateReward}{CalculateReward}
    \SetKwProg{Fn}{Function}{:}{}
    
    \Fn{\CalculateReward{$R^*, R^*_{prev}, R, R_{prev}, e, \eta$}}{
        \tcp{Calculate Global and Local Improvements}
        $\Delta \text{Global} \leftarrow \max \left( \frac{R^* - R^*_{prev}}{R^*_{prev}}, 0 \right)$ \;
        $\Delta \text{Local} \leftarrow \max \left( \frac{R - R_{prev}}{R_{prev}}, 0 \right)$ \;      
        
        \tcp{Calculate Weighted Improvement}
        $\Delta \text{Improvement} \leftarrow \Delta \text{Global} \cdot \eta + \Delta \text{Local} \cdot (1 - \eta)$ \;
        $\text{OC} \leftarrow \max (\Delta \text{Improvement})$ \;          
        
        \tcp{Check for Improvement}
        \If{$\Delta \text{Global} > 0 \text{ or } \Delta \text{Local} > 0$}{
            $r \leftarrow \frac{\Delta \text{Improvement} \times e}{C}$ \;
        }
        \Else{
            $r \leftarrow \frac{(\Delta \text{Improvement} - \text{OC}) \times e}{C}$ \tcp*{Include opportunity cost if no improvement}
        }
        
        \Return{$r$}
    }
\end{algorithm}

\section{Experimental design}
\label{sec5}
\subsection{Experimental setting}

The study scenario in this research is based on an unmanned warehouse $X$, located in Linyi City, Shandong Province, China. This warehouse primarily serves to consolidate and store regional less-than-truckload (LTL) cargo. Efficient warehouse operations, especially during inbound and outbound processes, are critical. Warehouse $X$ features docks on both sides for loading and unloading. Storage positions within the warehouse are allocated based on proximity to these docks, with the goal of storing goods as close as possible to the loading and unloading areas.  Although Warehouse $X$ has an I-shaped layout, its operational flow more closely resembles a U-shape, making it well-suited for using MSM docks \citep{rijal_integrated_2019}. In this layout and service mode, the docks at both ends are not designated exclusively for unloading or loading; their functions are determined based on the demand structure, which aligns well with on-site management needs.

\begin{figure}[ht] 
\centering
\includegraphics[scale=0.4]{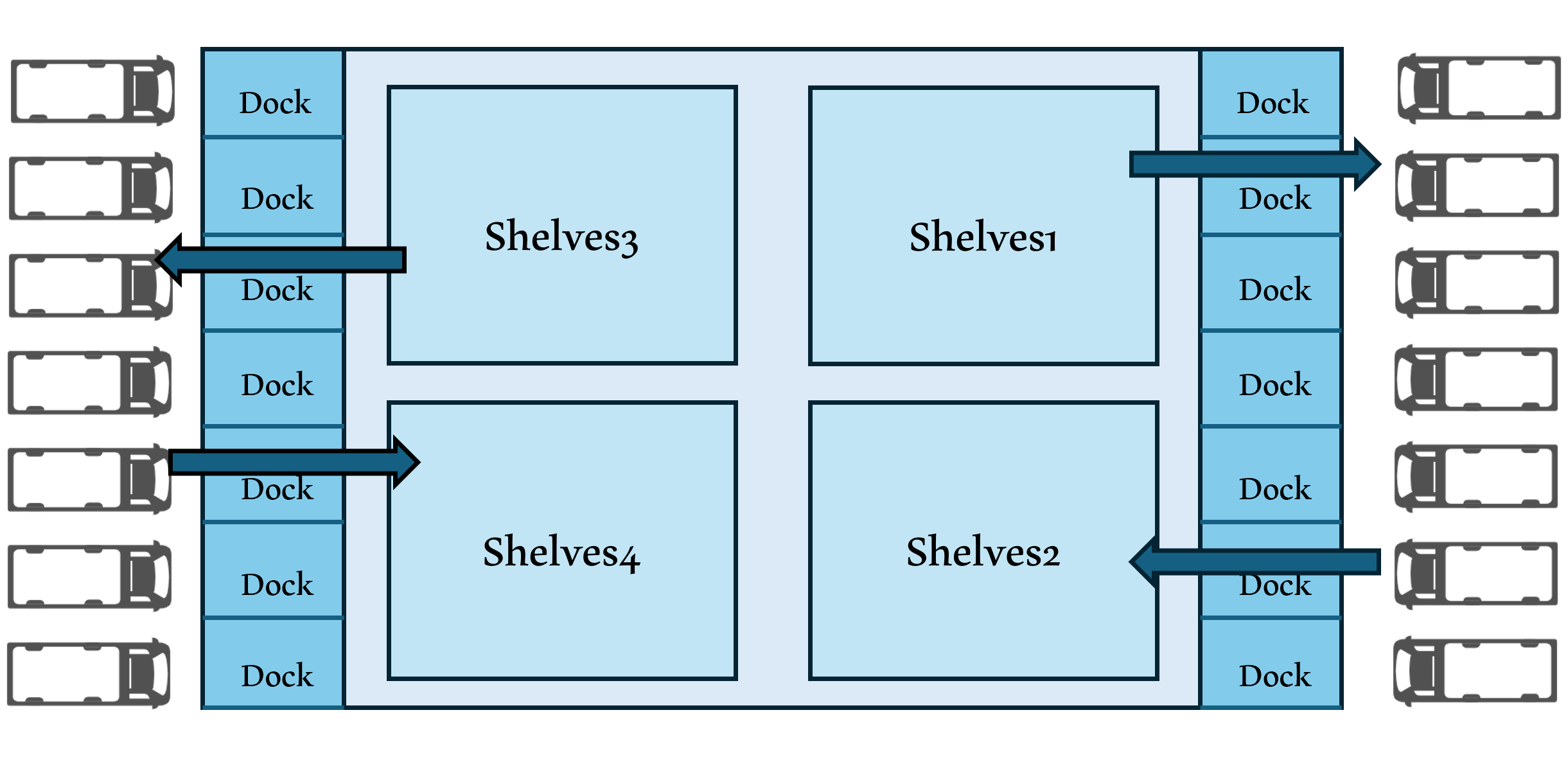} 
\caption{Layout and typical operational flow of warehouse $X$} 
\label{fig:layout} 
\end{figure}

To verify the effectiveness of the proposed algorithm, we apply it to a real-world scenario involving seven docks, using truck arrival and departure data along with cargo types collected from surveys.  By adjusting the scale of docks and trucks, we form ten sets of test instances, as shown in Table~\ref{table:test_instances}. Each truck has a predefined type, arrival and expected departure time window, total cargo volume, and the quantity of each cargo type. The positions of the docks and shelves are represented by coordinates, with the corresponding shelf number for each cargo type known. Following this setup,  we design a three-phase experiment. 

\begin{table}[ht]
\centering
\scriptsize
\caption{Set of test instances for TASP-DMD}
\label{table:test_instances}
\begin{tabularx}{\textwidth}{c>{\centering\arraybackslash}X>{\centering\arraybackslash}X|c>{\centering\arraybackslash}X>{\centering\arraybackslash}X}
\toprule
Instance & Number of Docks & Number of Trucks  & Instance & Number of Docks & Number of Trucks \\
\midrule
1 & 3 & 20 & 6 & 8 & 100 \\
2 & 4 & 30 &  7 & 8 & 130  \\
3 & 5 & 50 &  8 & 9 & 160  \\
4 & 6 & 60 &  9 & 9 & 180 \\
5 & 7 & 70 &   10 & 10 & 200  \\
\bottomrule
\end{tabularx}
\end{table}

In Phase 1, we conduct the operator filtering (OF) experiments to identify better-performing operator combinations. The goal is to pinpoint the operators that exhibit superior performance. In the Q-ALNS, the outer loop consists of several perturbation operators, and the inner loop includes a set of local search operators. Given our two-layer loop structure, the experiment is designed to first compare the performance of local search operators, addressing the question, ``Which local search operators perform better?'' After this, we combine the better-performing local search operators with perturbation operators to determine ``Which perturbation operators perform better?''

In Phase 2, we use a set of benchmarks to validate and evaluate the competitiveness of the proposed Q-ALNS, as outlined in Table~\ref{table:phase 3}. ALNS\_$i$ refers to the original ALNS, with $i$ indicating different combinations of operators based on the number of better-performing operators identified. RLNS replaces the AOS mechanism in the ALNS framework with a random selection mechanism. S-ALNS is a commonly used score-based ALNS in literature \citep{bodnar_scheduling_2017,rijal_integrated_2019}. Gurobi 11.0.1 is also included as a commercial solver for comparison. The evaluation comprehensively covers multiple aspects: efficiency and accuracy through comparison with benchmarks, multi-objective optimization capabilities via analysis of Pareto fronts, and adaptability to various problem instances. 
We also examine the impact of different parameter settings, action set size, and reward functions on the algorithm's performance. These analyses ensure a thorough understanding of the strengths of the Q-ALNS.

\begin{table}[htbp]
\centering
\scriptsize
\caption{Comparison benchmarks against the Q-ALNS}
\label{table:phase 3}
\begin{tabularx}{\textwidth}{cc>{\centering\arraybackslash}X}
\toprule
Index & Benchmark algorithm & Comparison purpose \\
\midrule
1 & ALNS\_$i$ &  Effectiveness of multiple operator combinations against a single operator combination \\
2 & RLNS & Effectiveness of Q-learning-based AOS \\
3 & S-ALNS & Effectiveness of Q-learning-based AOS \\
4 & Gurobi & Validate and Evaluate multi-objective optimization capabilities \\
\bottomrule
\end{tabularx}
\end{table}

In Phase 3, we analyze different dock mode strategies to demonstrate the effectiveness of the improvements in our model.
 Previous studies have explored several common dock operation strategies, including exclusive mode (fixed loading and unloading dock assignments), partial mix mode (pre-setting some MSM docks), and fully mixed mode (pre-setting all docks as MSM docks). To achieve greater flexibility, this paper introduces an adaptive dock mode decision strategy. To validate the effectiveness of this strategy, we compare it with the exclusive mode and the fully mixed mode. Since the partial mix mode closely resembles the strategy proposed in this paper, it is excluded from the comparison. We refer to the three strategies compared as Adaptive (this paper), Fix, and Mix strategies.

\subsection{Performance comparison metrics}

Each experiment, involving various instances, operators, and mechanisms, was independently run 10 times. For each run, we recorded the mean, the best value found for every objective, and CPU time. The comparison metrics used in the experiment are shown in Table~\ref{table:comparison_metrics}.

\begin{table}[htbp]
\centering
\scriptsize
\caption{Comparison Metrics used in the experiments}
\label{table:comparison_metrics}
\begin{tabularx}{\textwidth}{c p{8cm} c>{\centering\arraybackslash}X}
\toprule
Phase & Metric & Index & Purpose \\
\midrule
1 & Dominate ratio & \(Z_i\) & Quality \\
2 & Relative percentage deviation & RPD & Accuracy \\
2 & Time taken to reach the best solution & CPU time & Efficiency \\
2 & Nondominance Ratio & NR & Quality \\
2 & Hypervolume & HV & Diversity \\
2 & Hierarchical cluster counting & HCC & Uniformity \\
3 & Relative percentage deviation & RPD & Accuracy \\
3 & Time taken to reach the best solution & CPU time & Efficiency \\
\bottomrule
\end{tabularx}
\end{table}

In Phase 1, we used a dominance ratio metric to quantify the performance advantage of each local search and perturbation operator in our experiments. The dominance ratio \(Z_i\) for a given operator \(op_i\) is defined as the proportion of pairwise comparisons where \(op_i\) shows a statistically significant performance improvement over other operators. The result of this test is denoted as \(\zeta_{i,j}\), where:

\small
\begin{equation}
\zeta_{i,j} =
\begin{cases} 
1 & \text{if } op_i \text{ significantly outperforms } op_j, \\
-1 & \text{if } op_i \text{ is significantly outperformed by } op_j, \\
0 & \text{if there is no significant difference between } op_i \text{ and } op_j.
\end{cases}
\end{equation}

Next, we use \eqnref{eq:count} to count the number of times each operator \(op_i\) wins in the comparison. Here, \(n\) represents the total number of operators, and \(\delta(\cdot)\) is the indicator function, which equals 1 if the condition inside is true and 0 otherwise. The dominance ratio \(Z_i\) is then calculated as the ratio of the count \(C_i\) to the total number of pairwise comparisons, using \eqnref{eq:ratio}.

\small
\begin{equation}
   \ C_i = \sum_{j=1, j \neq i}^{n} \delta(\zeta_{i,j} = 1)
\label{eq:count}
\end{equation}

\small
\begin{equation}
    Z_i = \frac{C_i}{n}
\label{eq:ratio}
\end{equation}

In Phase 2, we evaluate the quality of the solutions using two types of methods. We first focus on three key metrics: relative percentage deviation (RPD) of the mean and best values, and the average computational time (CPU time) required to obtain the best solution. Lower values in these metrics indicate better performance of operators and selection mechanisms. As the tardiness objective is a shared concern for both trucks and warehouses, the RPD calculation is based on the tardiness. The formula for RPD is provided in \eqnref{RPD}, where \(R_i\) represents the tardiness value of experiment \(i\), and \(R_{\text{best}}\) is the best objective value (tardiness) found among all experiments.

\small
\begin{equation}
\label{RPD}
\text{RPD}_i = \frac{(R_i - R_{\text{best}})}{R_{\text{best}}} \times 100\%
\end{equation}

To evaluate the ability to find Pareto fronts, we employ three metrics: nondomination ratio (NR) \citep{tan_evolutionary_2001}, hypervolume (HV) \citep{zitzler_multiobjective_1999}, and hierarchical cluster counting (HCC) \citep{guimaraes_quality_2009}. The NR measures the proportion of solutions in the combined Pareto front that are non-dominated. Let \( P_B \) and \( P_C \) be the Pareto solution sets obtained by two different algorithms \( B \) and \( C \), respectively. The NR is calculated as \eqnref{NR}, where \( | \cdot | \) denotes the cardinality of the set. A higher NR value indicates that algorithm \( B \) contributes more useful candidate solutions to the combined Pareto front, reflecting its superior performance. 

\small
\begin{equation}
\label{NR}
NR_B = \frac{|\{ \mathbf{x} \in P_B \, | \, \mathbf{x} \text{ is nondominated in } P_{B \cup C}|}{|P_{B \cup C}|}
\end{equation}

The formula for HV is provided in \eqnref{HV}, where \(S\) represents the set of Pareto solutions, \((f_1, \ldots, f_m)\) are the objective functions, and \((f_1^*, \ldots, f_m^*)\) is the reference point. This metric calculates the volume of the objective space that is dominated by the Pareto front and bounded by the reference point. It provides a measure of both the convergence and diversity of the solutions. 

\small
\begin{equation}
\label{HV}
         HV(S) = \text{volume} \left( \bigcup_{x \in S} [f_1(x), f_1^*] \times \cdots \times [f_m(x), f_m^*] \right)
\end{equation}

The HCC metric assesses the diversity of solutions by performing hierarchical clustering and counting the resulting clusters. The process involves applying a clustering algorithm to group the solutions and then summing the distances between these clusters. The formula for HCC is provided in \eqnref{HCC}, where \( k \) represents the number of clusters and \( d_i \) denotes the distances between them. This metric offers insights into the distribution and variety of solutions within the Pareto front.

\small
\begin{equation}
\label{HCC}
HCC = \sum_{i=1}^{k-1} d_i
\end{equation}

In Phase 3,  we continue to use RPD and CPU time as metrics for comparison.
Furthermore, all statistical comparisons are conducted using the Wilcoxon signed-rank test \citep{wilcoxon_individual_1992}.
This step is essential to ensure the stability and reliability of the results across different instances and run times, enabling us to validate whether the observed differences in performance are statistically significant. This analysis confirms the robustness of the proposed algorithm across various scenarios.

\subsection{Parameters tuning}
\label{parameter tuning}

Our algorithm includes eight key parameters: \(\tau\), \(\epsilon\), \(\beta\), \(\alpha\), \(\gamma\), \(t\), \(\eta\), and \(l\), as shown in Table~\ref{parameter}. 
For parameter tuning, we employ response surface methodology (RSM). We begin by applying the Box-Behnken design method to create 112 parameter combinations. These combinations were tested on 10 instances, each run 5 times, with a stopping criterion of 400 outer loops. This criterion was based on the convergence trends observed in our experiments, which indicated that approximately 400 generations represent the overall trend. We then determine the optimal parameter combination using RSM and Pareto optimization. A detailed analysis of each parameter's influence on performance is provided in Section~\ref{sec6}.

\begin{table}[ht]
\centering
\caption{Parameters of the Q-ALNS, their corresponding levels, and tuned values}
\scriptsize
\begin{tabular}{cccccc}
\hline
\multirow{2}{*}{Parameter}      & \multirow{2}{*}{Notation} & \multicolumn{3}{c}{Levels} & \multirow{2}{*}{Tuned value} \\
                                &                           & -1      & 0        & 1     &                              \\ \hline
Temperature scale               & $\tau$                       & 0.01     & 0.105     & 0.2     &                              0.2\\
Epsilon-greedy                  & $\epsilon$                   & 0.7     & 0.85     & 1     &                              1\\
Epsilon-decay                   & $\beta$                      & 0.99    & 0.995    & 1     &                              0.99\\
Learning rate                   & $\alpha$                     & 0.5     & 0.75     & 1     &                              0.5\\
Discount factor                 & $\gamma$                      & 0.7     & 0.85     & 1     &                              0.7\\
Non\_improve times              & $t$                         & 10      & 25       & 40    &                              20\\
Local/global improvement weight & $\eta$                       & 0.6     & 0.7      & 0.8   &                              0.8\\ 
Learning loop                    & $l$                         & 100      & 200       & 300    &                              200\\\hline
\end{tabular}
\label{parameter}
\end{table}

\section{Numerical and statistical results}
\label{sec6}

This section executes the three-phase experiment outlined in Section~\ref{sec5} to validate the effectiveness and competitiveness of the Q-ALNS. All experiments were conducted using Python 3.8 on the Snellius Cluster High-Performance Computer, specifically on an AMD Rome 7H12 with CPUs running at 2.6 GHz with 32GB of RAM. Unless otherwise stated, the comparison results have passed significance tests.

\subsection{Phase 1: Comparison between operators}
\label{phase1}

In this phase, we identify the most advantageous operators for the problem through a series of experiments, assessing their impact on search quality and efficiency.

Common ranking methods in the literature for comparing operator performance include frequency-based, ablation-based, and pairwise comparison matrices. Given the multiple operators in the Q-ALNS, we employ the pairwise comparison metric method using the Wilcoxon test. This approach involves comparing operators in pairs and filling a pairwise comparison matrix based on statistically significant performance differences, resulting in a dominance matrix for all operators.

The local search operators in the Q-ALNS consist of destroy and repair operators, as listed in Table~\ref{tab:operation}. These operators are selectively combined into specific combinations, each tailored to the objective functions (see Table~\ref{tab:combination}). The sources and detailed descriptions are provided in ~\ref{app2}. For the perturbation phase of the algorithm, we design three types of perturbation operators tailored for unloading-only (I), loading-only (O), and mixed-mode (F) docks, as shown in Table~\ref{tab:perturbation}.

\begin{table}[ht]
\centering
\scriptsize
\caption{Description of local search operators}
\label{tab:operation}
\begin{tabularx}{\textwidth}{cc>{\centering\arraybackslash}X}
\toprule
ID & Index & Destroy or repair operation \\
\midrule
1  & rRd         & Randomly remove a truck from all available trucks \\
2  & rMxTar      & Remove the truck causing the most significant delay \\
3  & rMxM        & Remove the truck with the greatest weighted cargo handling distance \\
4  & iBck        & Swap the removed truck with its preceding truck at the same dock \\
5  & iFwd        & Swap the removed truck with its succeeding truck at the same dock \\
6  & iSwap       & Swap the removed truck with another randomly selected unloading or loading truck \\
7  & iUp         & Reallocate the removed truck to a dock with a higher rank based on handling distance \\
8  & iDown       & Reallocate the removed truck to a dock with a lower rank based on handling distance \\
9  & iDockInsert & Insert the removed truck at all positions in the truck sequence of the same dock \\
10 & iBtwInsert  & Insert the removed truck randomly in the truck sequence of different docks \\
11 & riInD2D     & Swap trucks between two unloading docks \\
12 & riOuD2D     & Swap trucks between two loading docks \\
13 & riFlxD2D    & Swap trucks between two mixed-mode docks \\
\bottomrule
\end{tabularx}
\end{table}

\begin{table}[H]
\centering
\scriptsize
\caption{The combination of local search operators used in the Q-ALNS}
\label{tab:combination}
\begin{tabularx}{\textwidth}{cc>{\centering\arraybackslash}X|cc>{\centering\arraybackslash}X} 
\toprule
ID & Operators & Target & ID & Operators & Target \\
\midrule
$op_1$  & rRd \& iBck        & Sequence Adjustment    & $op_9$  & rMxM \& iUp        & Sequence \&Dock Adjustment \\
$op_2$  & rRd \& iFwd        & Sequence Adjustment    & $op_{10}$ & rRd \& iDockInsert & Sequence \&Dock Adjustment \\
$op_3$  & rRd \& iSwap       & Sequence \&Dock Adjustment & $op_{11}$ & rRd \& iBtwInsert  & Sequence \&Dock Adjustment \\
$op_4$  & rRd \& iUp         & Dock Adjustment        & $op_{12}$ & rMxTar \& iDockInsert & Sequence \&Dock Adjustment \\
$op_5$  & rRd \& iDown       & Dock Adjustment        & $op_{13}$ & rMxTar \& iBtwInsert  & Sequence \&Dock Adjustment \\
$op_6$  & rMxTar \& iBck     & Sequence Adjustment    & $op_{14}$ & riInD2D            & Dock Adjustment    \\
$op_7$  & rMxTar \& iSwap    & Distance \&Dock Adjustment & $op_{15}$ & riOuD2D            & Dock Adjustment    \\
$op_8$  & rMxM \& iSwap      & Distance \&Dock Adjustment & $op_{16}$ & riFlxD2D           & Dock Adjustment    \\
\bottomrule
\end{tabularx}
\end{table}

\begin{table}[ht]
\centering
\scriptsize
\caption{Description of perturbation operators}
\label{tab:perturbation}
\begin{tabularx}{\textwidth}{
    m{0.1\textwidth} 
    m{0.15\textwidth} 
    >{\centering\arraybackslash}X 
}
\toprule
ID & Index & Perturbation operation \\
\midrule
$P_1$   & Chn2I & Randomly selects a dock and changes its type to unloading-only (I) \\
$P_2$   & Chn2F & Randomly selects a dock and changes its type to mixed-mode (F) \\
$P_3$   & Chn2O & Randomly selects a dock and changes its type to loading-only (O) \\
\bottomrule
\end{tabularx}
\end{table}

\begin{figure}[ht] 
\centering
\includegraphics[scale=0.5]{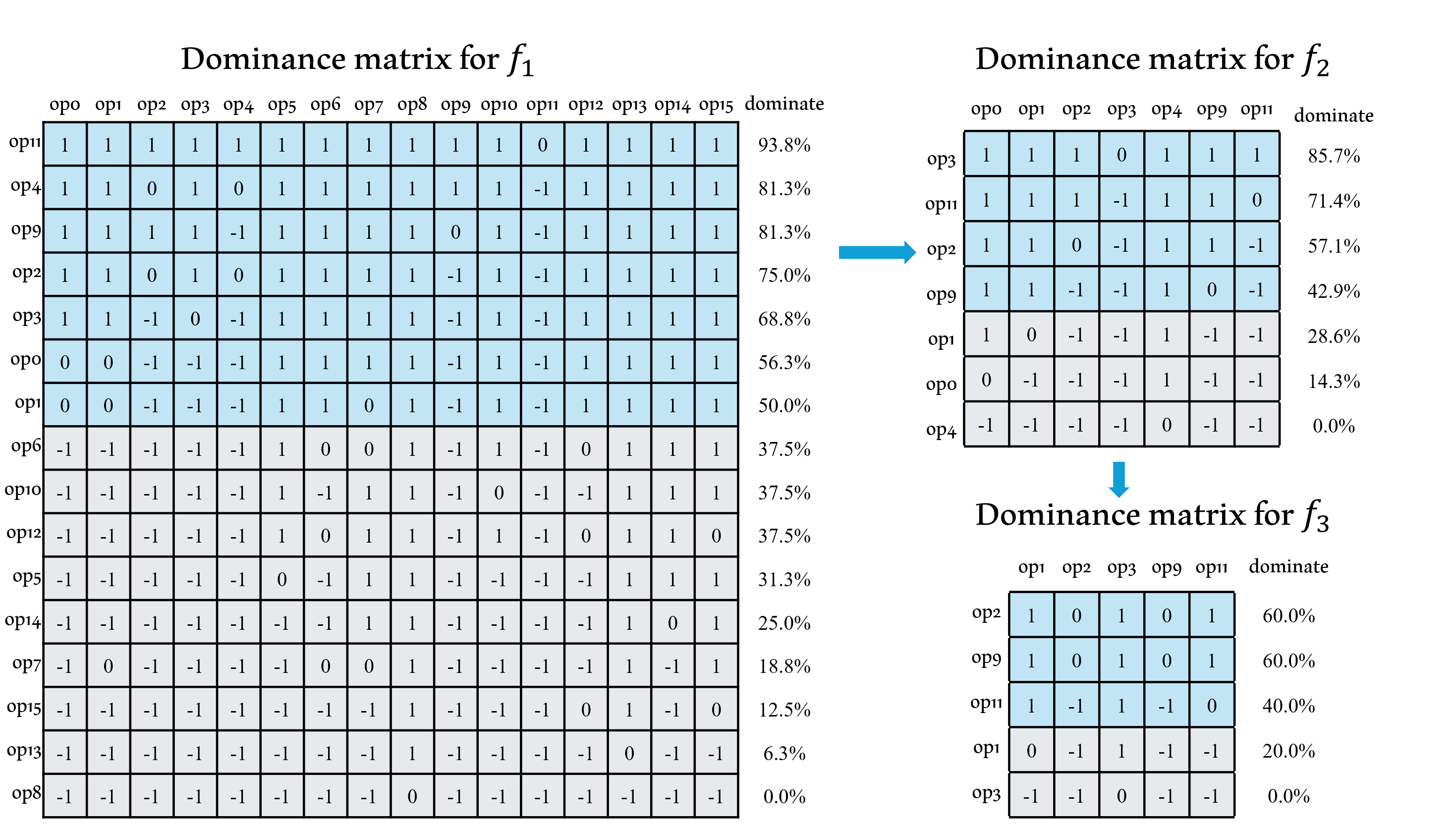} 
\caption{Local search operator filtering based on the Wilcoxon dominance matrix} 
\label{fig:dominate matrix} 
\end{figure}

In the OF experiments, we evaluate the performance of local search operators in the inner loop. Each local search operator \(op_i\) is independently tested over 10 instances with 100 iterations each, and each experiment is repeated 10 times to ensure reliability. We set a dominance ratio threshold of 40\%  to identify better-performing local search operators. Operators that outperform 40\% of others in pairwise comparisons are considered superior. With three prioritized objective functions  \(f_1 > f_2 > f_3\), operators are filtered accordingly. The filtering process, illustrated in Figure~\ref{fig:dominate matrix}, identify operators \(op2\), \(op9\), and \(op11\) as the better-performing local search operators. Additionally, we compare the performance of perturbation operators in the outer loop over 30 iterations and find that $P_1$ significantly outperforms $P_0$ and $P_2$.

\begin{figure}[H] 
\centering
\includegraphics[scale=0.3]{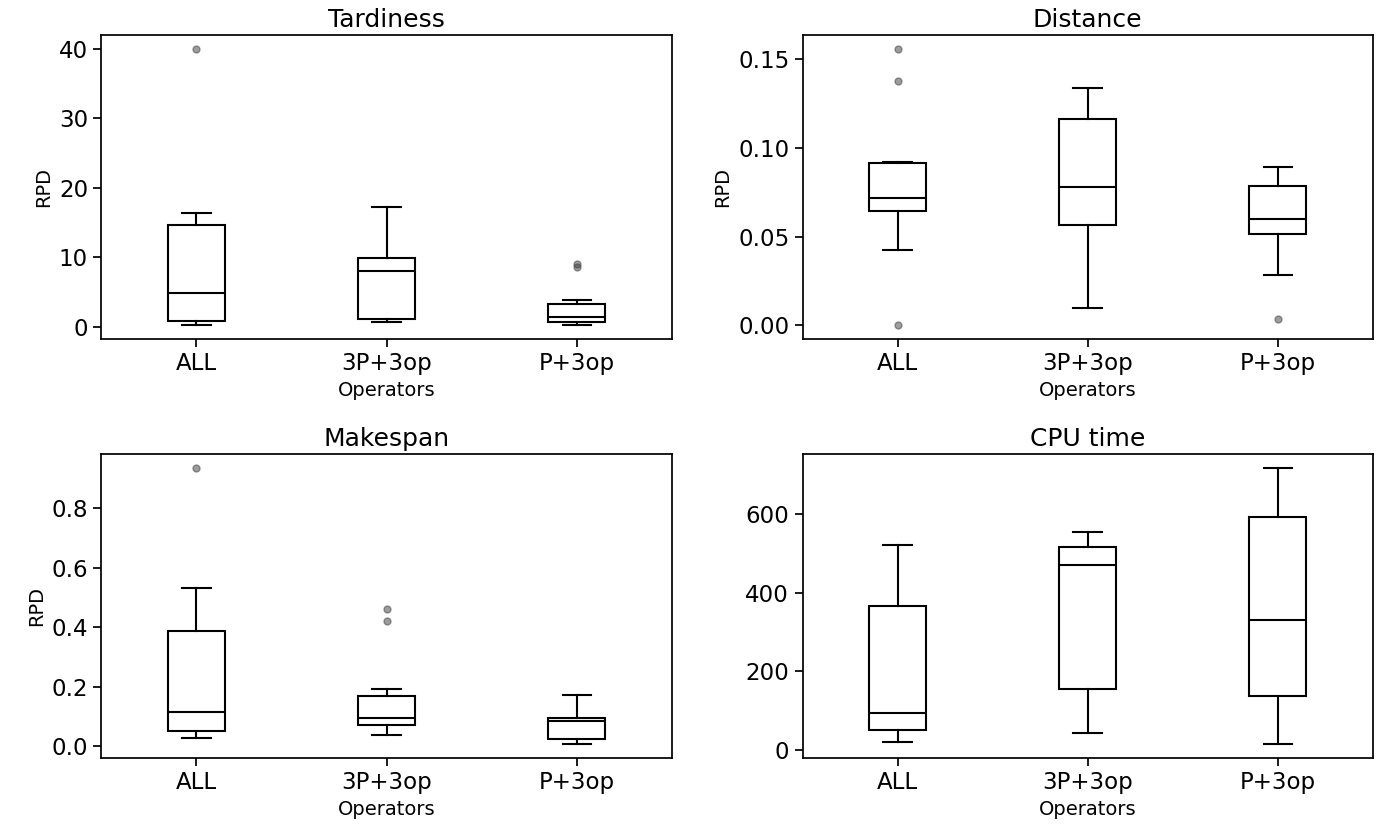} 
\caption{Comparison of three types of operators sets for the instance with 7 docks and 70 trucks} 
\label{fig:operator_filter} 
\end{figure}

From the OF experiments, we reduce the original set of 16 local search operators to 3 and identify a single effective perturbation operator. Figure~\ref{fig:operator_filter} shows the performance comparison among three sets of operators: all operators (ALL), all perturbation operators combined with the three better-performing local search operators ('3P+3op'), and the better-performing perturbation operator combined with the three better-performing local search operators ('P+3op'). The 'P+3op' set achieves better values across all objective functions and demonstrates higher stability, with a moderate increase in CPU time.

\subsection{Phase 2: Performance analysis of the Q-ALNS against benchmark algorithms}

In this stage of the experiment, we first compare three ALNS\_$i$ mechanisms with RLNS to assess the impact of better-performing operator combinations on the exploration-exploitation capability of the ALNS. Next, we compare the Q-ALNS with RLNS and the commonly used S-ALNS to demonstrate the effectiveness of the Q-learning-based operator selection mechanism compared to random and score-based selection mechanisms.

\subsubsection{Comparison based on optimality gap and CPU time }

Table~\ref{tab:overall_performance} presents the performance of three ALNS\_$i$ algorithms, RLNS, S-ALNS, and the Q-ALNS, across 10 instances. Since we filtered three local search operators, the ALNS\_$i$ algorithm has three versions: ALNS\_$1$, ALNS\_$2$, and ALNS\_$3$. The ``AvS'' column shows the average RPD (ARPD) of each algorithm from 10 independent runs per instance, while the ``BS'' column displays the RPD of the best solution. The ``T(s)'' column indicates the average CPU time. Figure~\ref{fig:boxplot_6_algorithms_td} shows the boxplots of RPD for tardiness and CPU time, respectively.

Comparison of ALNS\_$1$, ALNS\_$2$, ALNS\_$3$, and RLNS data shows RLNS performs better in both ``AvS'' and ``BS'' (lower values). Figures~\ref{fig:boxplot_6_algorithms_td} shows RLNS with a lower mean (red dots) and minimum value, and a smaller interquartile range, indicating better robustness. This indicates that the use of multiple operators, as seen in RLNS, significantly enhances the competitiveness of ALNS in solving the TASP-DMD problem, compared to relying on individual operators. While all algorithms utilize the best operators, RLNS benefits from combining multiple operators, leading to improved performance.

The Q-ALNS achieves better AvS and BS values across all instances, particularly excelling in large-scale cases. It consistently performs well across different instance scales. The ``Average'' row in Table~\ref{tab:overall_performance} indicates that Q-learning improves ARPD and best RPD by 9.0\% and 52.9\%, respectively, compared to RLNS, and by 9.4\% and 49.8\% compared to S-ALNS.
According to the average CPU time reported in Table~\ref{tab:overall_performance} and the CPU time boxplot in Figure~\ref{fig:boxplot_6_algorithms_td}, the additional mechanisms of Q-ALNS introduce some overhead complexity. The Q-ALNS initially shows the longest CPU time, especially for smaller instances. However, as instance sizes increase, the relative CPU time for it decreases, and it is no longer the highest. This change reflects better scalability and efficiency in larger problem sizes. 

We further compared the convergence behavior of the algorithms. Figure~\ref{fig:converge_instance_compare} shows the convergence curves of all algorithms for the representative cases (3\_20 and 10\_200). The Q-ALNS demonstrates a lower convergence curve with consistent behavior during the search process. It shows faster initial convergence, as evidenced by the rapid decline in its convergence curve during early iterations, and achieves lower final convergence values, reflecting its capability to find better final solutions. The convergence curves of Q-ALNS are smoother and exhibit less fluctuation, suggesting a more stable search process and reduced interference from local optima. 

\begin{landscape}
\begin{table}[htbp]
\centering

\caption{Overall comparison of the Q-ALNS and classical ALNS}
\label{tab:overall_performance}
\scriptsize
\begin{tabular}{cc|ccc|ccc|ccc|ccc|ccc|ccc}
\toprule
\multicolumn{2}{c|}{Instance set} & \multicolumn{3}{c|}{ALNS\_$1$} & \multicolumn{3}{c|}{ALNS\_$2$} & \multicolumn{3}{c|}{ALNS\_$3$} & \multicolumn{3}{c|}{RLNS} & \multicolumn{3}{c|}{S-ALNS} & \multicolumn{3}{c}{Q-ALNS} \\
\cmidrule{1-2} \cmidrule{3-5} \cmidrule{6-8} \cmidrule{9-11} \cmidrule{12-14} \cmidrule{15-17} \cmidrule{18-20}
Dock & Truck & AvS & BS & T(s) & AvS & BS & T(s) & AvS & BS & T(s) & AvS & BS & T(s) & AvS & BS & T(s) & AvS & BS & T(s) \\
\midrule
3 & 20 & 2.749 & 2.618 & 4.2 & 1.939 & 1.875 & 24.6 & 0.624 & 0.320 & 43.8 & 0.290 & 0.155 & 25.3 & 0.258 & 0.151 & 25.8 & \cellcolor{gray!30}\textbf{0.256} & \cellcolor{gray!30}\textbf{0.008} & 28.7 \\
4 & 30 & 0.479 & 0.419 & 6.0 & 0.629 & 0.629 & 41.6 & 0.461 & 0.413 & 49.4 & 0.183 & 0.113 & 49.6 & \cellcolor{gray!30}\textbf{0.163} & 0.097 & 49.4 & 0.179 & \cellcolor{gray!30}\textbf{0.080} & 40.4 \\
5 & 50 & 0.344 & 0.273 & 9.7 & 0.509 & 0.473 & 82.4 & 0.337 & 0.226 & 108.0 & 0.085 & 0.029 & 108.8 & 0.076 & \cellcolor{gray!30}\textbf{0.004} & 108.0 & \cellcolor{gray!30}\textbf{0.065} & 0.023 & 118.9 \\
6 & 60 & 0.087 & 0.042 & 11.6 & 0.232 & 0.221 & 99.4 & 0.221 & 0.151 & 132.3 & 0.094 & \cellcolor{gray!30}\textbf{0.028} & 135.0 & 0.095 & 0.059 & 132.3 & \cellcolor{gray!30}\textbf{0.075} & 0.029 & 154.1 \\
7 & 70 & 8.795 & 8.622 & 13.6 & 9.322 & 9.095 & 133.2 & 1.015 & 0.576 & 560.9 & 0.386 & 0.218 & 247.6 & 0.446 & 0.198 & 254.5 & \cellcolor{gray!30}\textbf{0.367} & \cellcolor{gray!30}\textbf{0.106} & 216.4 \\
8 & 100 & 0.196 & 0.180 & 19.8 & 0.308 & 0.302 & 231.8 & 0.086 & 0.060 & 940.9 & 0.030 & 0.016 & 424.0 & \cellcolor{gray!30}\textbf{0.028} & 0.013 & 440.7 & 0.030 & \cellcolor{gray!30}\textbf{0.012} & 412.2 \\
8 & 130 & 0.057 & 0.045 & 26.3 & 0.144 & 0.135 & 363.0 & 0.062 & 0.049 & 1585.7 & 0.029 & 0.020 & 700.1 & \cellcolor{gray!30}\textbf{0.028} & 0.016 & 712.3 & 0.030 & \cellcolor{gray!30}\textbf{0.016} & 702.2 \\
9 & 160 & 0.063 & 0.055 & 32.7 & 0.141 & 0.138 & 498.5 & 0.031 & 0.024 & 2652.2 & \cellcolor{gray!30}\textbf{0.015} & 0.007 & 1149.4 & 0.019 & 0.011 & 1135.2 & \cellcolor{gray!30}\textbf{0.015} & \cellcolor{gray!30}\textbf{0.003} & 1128.4 \\
9 & 180 & 0.046 & 0.035 & 37.1 & 0.140 & 0.134 & 628.7 & 0.038 & 0.023 & 3418.4 & 0.021 & 0.014 & 1471.5 & 0.024 & 0.016 & 1545.9 & \cellcolor{gray!30}\textbf{0.015} & \cellcolor{gray!30}\textbf{0.006} & 1404.4 \\
10 & 200 & 0.041 & 0.037 & 42.8 & 0.140 & 0.137 & 707.9 & 0.027 & 0.020 & 5253.6 & 0.018 & 0.012 & 2172.7 & 0.018 & 0.009 & 2289.8 & \cellcolor{gray!30}\textbf{0.014} & \cellcolor{gray!30}\textbf{0.006} & 1973.5 \\
[1ex] 
\multicolumn{2}{c|}{Average} & 1.286 & 1.233 & 20.374 & 1.350 & 1.314 & 281.097 & 0.290 & 0.186 & 1505.232 & 0.115 & 0.061 & 648.402 & 0.115 & 0.057 & 669.385 & 0.105 & 0.029 & 617.6 \\
\bottomrule
\end{tabular}

\end{table}

\begin{figure}[H]
    \centering
    \begin{subfigure}{0.64\linewidth}  
        \centering
        \includegraphics[height=0.48\textheight]{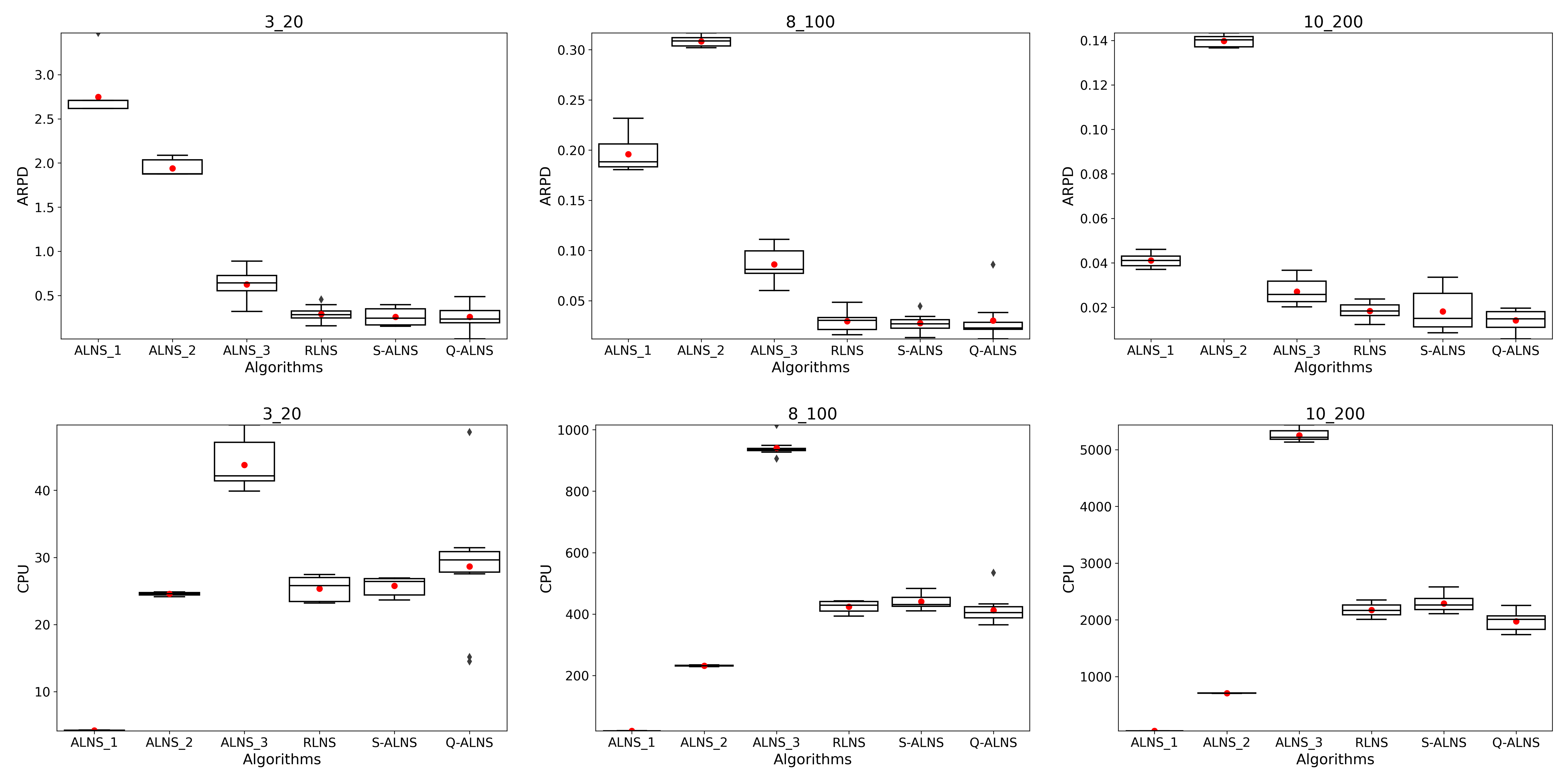}  
        \caption{Boxplot based on ARPD of tardiness and CPU}
        \label{fig:boxplot_6_algorithms_td}
    \end{subfigure}
    \hspace{0.05\linewidth}  
    \begin{subfigure}{0.3\linewidth}   
        \centering
        \includegraphics[height=0.48\textheight]{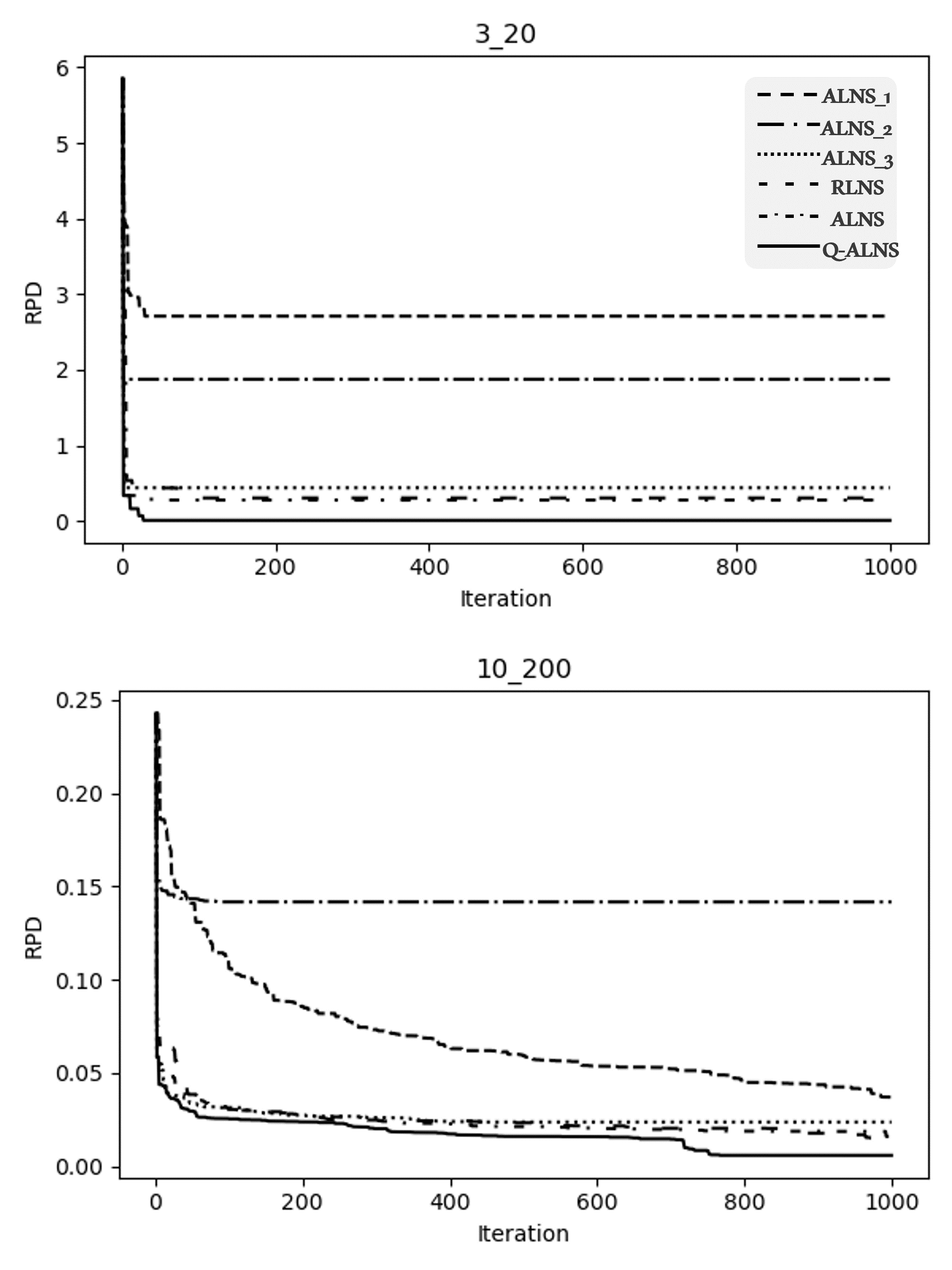}  
        \caption{Convergence rate}
        \label{fig:converge_instance_compare}
    \end{subfigure}
    \caption{Comparison of the Q-ALNS, RLNS, S-ALNS, and ALNS\_$i$ based on different metrics}
\end{figure}

\end{landscape}

This phase of experiments demonstrates the successful identification of better-performing operators suitable for the TASP-DMD problem. The integration of the Q-learning mechanism significantly enhances the algorithm's exploration of the solution space. While the Q-ALNS initially requires more computational time for smaller instances, its performance improves with larger problem sizes. This trade-off is justified by the significant improvements in solution quality and algorithm stability. These findings confirm that the Q-learning-based AOS mechanism not only improves solution quality but also ensures a more robust and efficient search process.

\subsubsection{Comparison based on Pareto front}

This section compares the Q-ALNS with Gurobi and RLNS to demonstrate its ability to find Pareto fronts.  Gurobi solutions, obtained using the $\epsilon$-constrains method, where  \( f_2 \) and \( f_3 \) are relaxed into constraints and solved with Gurobi 11.0.1 and Python 3.8, result in an approximate Pareto front.

Due to time constraints, Gurobi failed to find optimal solutions for most instances, so three small-scale instances were designed for comparison. 
Based on the data in Table~\ref{tab:validate_pareto_performance}, the Q-ALNS shows superior performance compared to Gurobi. It achieves a higher proportion of non-dominated solutions (NR) on average (51.9\%) compared to Gurobi (48.1\%), demonstrating better quality in finding Pareto-optimal solutions. It also outperforms Gurobi in hypervolume (HV), with an average HV of 0.638 compared to Gurobi's 0.577, indicating a more comprehensive exploration of the solution space. Expectedly, Gurobi performs better in hierarchical cluster counting (HCC), with an average HCC of 3.507 compared to Q-ALNS's 2.380, indicating better solution uniformity. Overall, the Q-ALNS is comparable to or better than Gurobi, validating its effectiveness in solving the TASP-DMD problem.

For larger-scale instances, the Q-ALNS is compared with RLNS, as shown in Table~\ref{tab:large_pareto_performance}. The Q-ALNS demonstrates superior performance over RLNS in most instances. It achieves a higher average NR value (0.653) compared to RLNS (0.347), indicating greater effectiveness in producing non-dominated solutions. In terms of HV, it averages 0.732, slightly outperforming RLNS's 0.668, although the \textit{p}-value of 0.375 indicates no significant difference. For HCC, the Q-ALNS performs better, with an average HCC of 15.011 compared to RLNS's 8.524, suggesting more evenly distributed solutions. 

\subsubsection{Adaptiveness of the Q-ALNS to problem instances}

This section analyzes how the Q-ALNS adaptively selects local search operators based on problem characteristics. We use the relative improvement index ($RI$) \citep{karimi-mamaghan_learning_2023} to evaluate the contribution of different local search operators throughout the search process. \( RI \) is calculated by  \eqnref{eq:RI}, where \( R(a) \) represents the average improvement per application of local search operator \( a \) over the entire search process. \( R(a) \) is determined by  \eqnref{eq:R}, where \( GI(a) \) represents the total global improvement contributed by operator $a$ and \( S(a) \) represents the number of improvements achieved by operator $a$. \( RI(a) \) then expresses this average improvement as a percentage of the total improvement achieved by all local search operators.

\begin{align}
&RI(a) = \frac{R(a)}{\sum_{a' \in A} R(a')} \times 100\% \label{eq:RI} \\
&R(a) = \frac{GI(a)}{S(a)} \label{eq:R}
\end{align}

\begin{figure}[H] 
\centering
\includegraphics[scale=0.4]{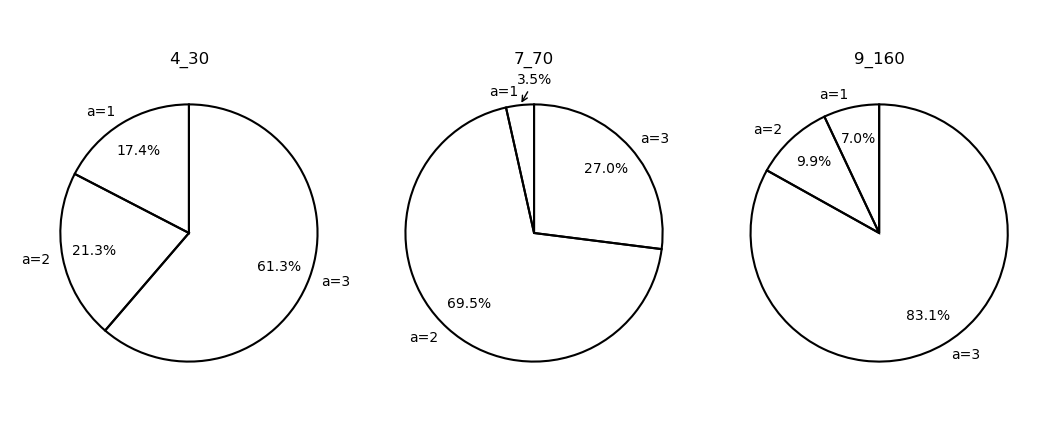} 
\caption{Average improvement performance of each local search operator in certain instances} 
\label{fig:RI} 
\end{figure}

Figure~\ref{fig:RI} shows the $RI$ results for three randomly selected instances with varying characteristics, including dock-to-truck ratios and problem sizes ranging from small to large. 
The pie charts illustrate the distribution of the RI for three different local search operators (\( a=1, a=2, a=3 \)) across the three instances (4\_30, 7\_70, and 9\_160). The results show clear variations in each operator's contribution depending on the instance characteristics. For example, in instance 4\_30, operator \( a=3 \) has the highest contribution at 61.3\%, while in instance 7\_70, operator \( a=2 \) dominates with 69.5\%. As problem size increases, such as in instances 9\_160, operator \( a=3 \) becomes increasingly dominant, contributing over 83\%. These results demonstrate the Q-ALNS's ability to adaptively select the most suitable local search operators according to problem features, efficiently handling a wide range of problem scales and complexities.

\begin{figure}[ht] 
\centering
\includegraphics[scale=0.5]{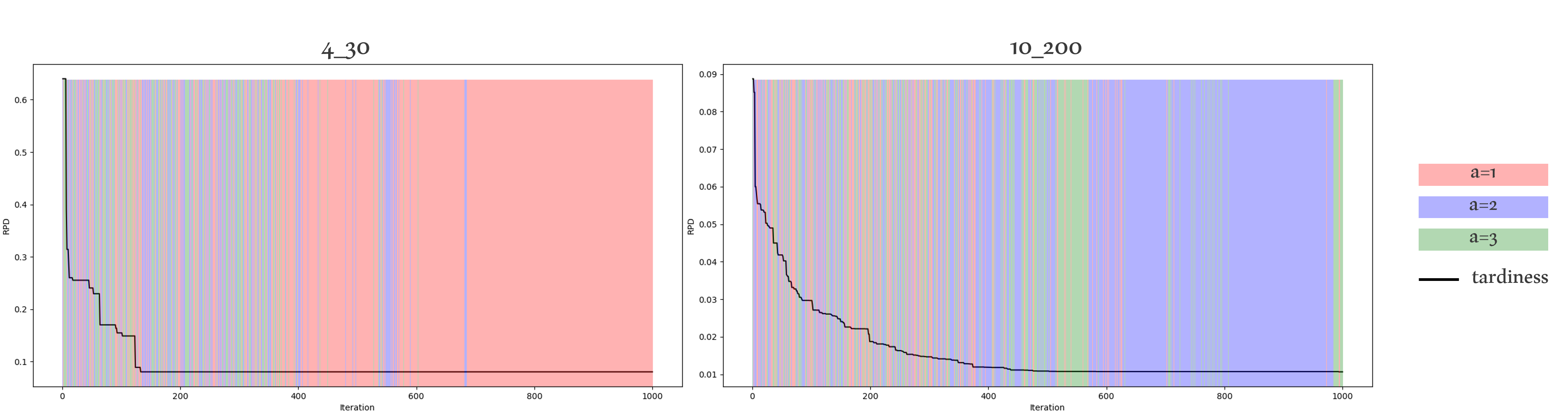} 
\caption{Selection of local search operators at each iteration in the search process for certain instances} 
\label{fig:selection step} 
\end{figure}

Figure~\ref{fig:selection step} provides a detailed visualization of operator selection throughout the search process for typical instances 4\_30 and 10\_200. Different colored bars represent the selected local search operator for each iteration (red for \( a=1 \), purple for \( a=2 \), and green for \( a=3 \)), while the black solid line represents the convergence curve. The charts show the dynamic selection of local search operators during the search process. In instance 4\_30, the algorithm predominantly uses operator \( a=1 \) (red), contributing significantly to rapid convergence. In instance 10\_200, as the problem size increases, operator \( a=3 \) (green) becomes crucial in the mid to late iteration. 
This dynamic adaptation of operator selection throughout the iterations underscores the Q-ALNS's ability to tailor its search strategy based on evolving problem characteristics, enhancing search efficiency and improving solution quality across different problem scales.

\begin{table}[htbp]
\centering
\scriptsize
\caption{Overall comparison of the Q-ALNS with Gurobi based on Pareto front}
\label{tab:validate_pareto_performance}
\begin{tabular}{c|c|c|c|c|c|c|c} 
\toprule
\multicolumn{2}{c|}{Instance} & \multicolumn{2}{c|}{NR} & \multicolumn{2}{c|}{HV} & \multicolumn{2}{c}{HCC} \\
\cmidrule{1-2} \cmidrule{3-4} \cmidrule{5-6} \cmidrule{7-8}
Dock & Truck & Gurobi & Q-ALNS & Gurobi & Q-ALNS & Gurobi & Q-ALNS \\
\midrule
6 & 20 & 30.0\% & \cellcolor{gray!30}\textbf{70.0\%} & \cellcolor{gray!30}\textbf{0.313} & 0.310 & \cellcolor{gray!30}\textbf{3.588} & 2.096 \\
7 & 20 & \cellcolor{gray!30}\textbf{50.0\%} &\cellcolor{gray!30}\textbf{50.0\%} & 0.721 &\cellcolor{gray!30}\textbf{0.768} &\cellcolor{gray!30}\textbf{3.254} & 2.535 \\
8 & 20 & 64.3\% & \cellcolor{gray!30}\textbf{35.7\%} & 0.698 & \cellcolor{gray!30}\textbf{0.835} & \cellcolor{gray!30}\textbf{3.679} & 2.509 \\
[1ex] 
\multicolumn{2}{c|}{Average} & 48.1\% & \cellcolor{gray!30}\textbf{51.9\%} & 0.577 & \cellcolor{gray!30}\textbf{0.638} & \cellcolor{gray!30}\textbf{3.507} & 2.380 \\
\bottomrule
\end{tabular}
\end{table}

\begin{table}[ht]
\centering
\scriptsize
\caption{Overall comparison of the Q-ALNS with RLNS based on Pareto front}
\scriptsize
\label{tab:large_pareto_performance}
\begin{tabularx}{\textwidth}{cc|XX|XX|XX}
\toprule
\multicolumn{2}{c|}{Instance} & \multicolumn{2}{c|}{NR} & \multicolumn{2}{c|}{HV} & \multicolumn{2}{c}{HCC} \\
\cmidrule(r){1-2} \cmidrule(r){3-4} \cmidrule(r){5-6} \cmidrule(r){7-8} 
Dock & Truck & RLNS & Q-ALNS & RLNS & Q-ALNS & RLNS & Q-ALNS \\
\midrule
3 & 20 & \cellcolor{gray!30}\textbf{0.648} & 0.352 & 0.775 & \cellcolor{gray!30}\textbf{0.827} & 5.670 & \cellcolor{gray!30}\textbf{8.363} \\
4 & 30 & 0.434 & \cellcolor{gray!30}\textbf{0.566} & \cellcolor{gray!30}\textbf{0.779} & 0.758 & 9.437 & \cellcolor{gray!30}\textbf{15.611} \\
5 & 50 & 0.269 & \cellcolor{gray!30}\textbf{0.731} & 0.502 & \cellcolor{gray!30}\textbf{0.709} & 7.743 & \cellcolor{gray!30}\textbf{14.079} \\
6 & 60 & 0.430 & \cellcolor{gray!30}\textbf{0.570} & \cellcolor{gray!30}\textbf{0.769} & 0.585 & 8.198 & \cellcolor{gray!30}\textbf{17.983} \\
7 & 70 & 0.484 & \cellcolor{gray!30}\textbf{0.516} & \cellcolor{gray!30}\textbf{0.915} & 0.775 & 9.280 & \cellcolor{gray!30}\textbf{23.192} \\
8 & 100 & 0.384 & \cellcolor{gray!30}\textbf{0.616} & \cellcolor{gray!30}\textbf{0.833} & 0.692 & 13.165 & \cellcolor{gray!30}\textbf{20.849} \\
8 & 130 & 0.308 & \cellcolor{gray!30}\textbf{0.692} & \cellcolor{gray!30}\textbf{0.734} & 0.682 & 6.937 & \cellcolor{gray!30}\textbf{14.113} \\
9 & 160 & 0.206 & \cellcolor{gray!30}\textbf{0.794} & 0.575 & \cellcolor{gray!30}\textbf{0.815} & 8.394 & \cellcolor{gray!30}\textbf{10.377} \\
9 & 180 & 0.111 & \cellcolor{gray!30}\textbf{0.889} & 0.458 & \cellcolor{gray!30}\textbf{0.712} & 7.301 & \cellcolor{gray!30}\textbf{12.064} \\
10 & 200 & 0.275 & \cellcolor{gray!30}\textbf{0.725} & 0.507 & \cellcolor{gray!30}\textbf{0.785} & 9.893 & \cellcolor{gray!30}\textbf{14.415} \\
[1ex] 
\multicolumn{2}{c|}{Average} & 0.347 & \cellcolor{gray!30}\textbf{0.653} & 0.668 & \cellcolor{gray!30}\textbf{0.732} & 8.524 & \cellcolor{gray!30}\textbf{15.011} \\
\bottomrule
\end{tabularx}
\end{table}

\subsubsection{Sensitivity analysis}

This section conducts a sensitivity analysis of the Q-ALNS's parameters, action space, and reward mechanism to assess how variations in these components affect the algorithm's performance and robustness. 

\subsubsection*{Q-ALNS parameters}

\begin{figure}[ht] 
\centering
\includegraphics[scale=0.4]{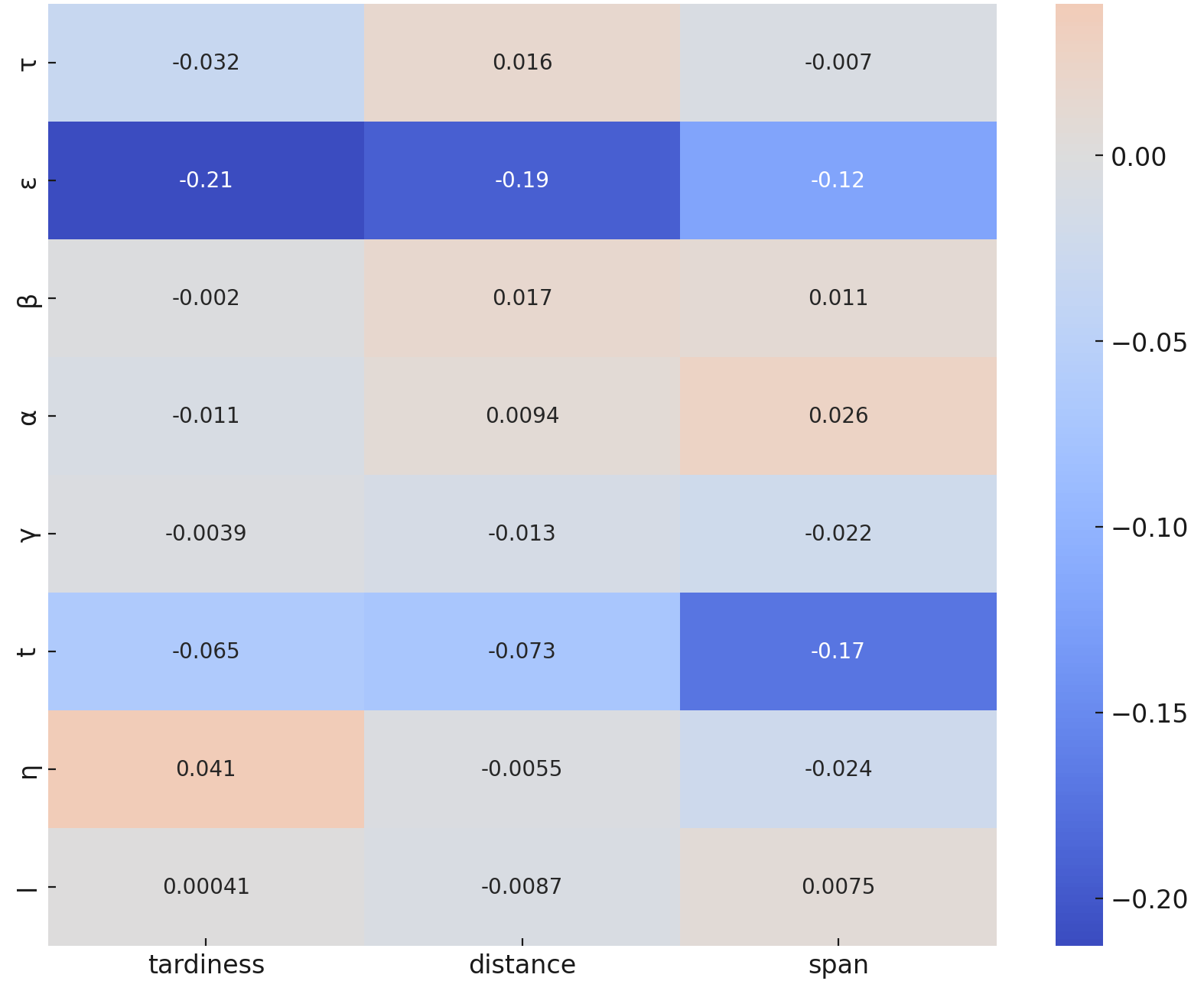} 
\caption{Correlation heatmap between parameters and objective functions} 
\label{fig:correltion} 
\end{figure}

The sensitivity analysis focuses on the impact of parameters on tardiness, distance, and makespan. The correlation heatmap (Figure~\ref{fig:correltion}) reveals that \(\epsilon\) and \(t\) have significant negative effects on all objectives, making them the most influential parameters. Other parameters show weaker and less consistent correlations. 
 To further investigate, we vary each parameter across its range while keeping others at optimal values, observing the changes in three objectives. The results, shown in Figure~\ref{fig:sensitive} with tuned values marked in Table~\ref{parameter}, reveal that the Q-ALNS is sensitive to parameter levels, which impact each objective differently.  Lower \(\epsilon\) values significantly reduce tardiness and makespan but increase distance. In contrast, $t$ shows a fluctuating impact on tardiness and distance, with a smaller effect on makespan. This highlights the importance of effective parameter tuning, as applied in Section~\ref{parameter tuning}.

\subsubsection*{Action set size}

The action space in our algorithm is determined by the number of local search operators. After selecting the better-performing local search operators, the Q-ALNS has an action set size of three. We compare this with a larger action space of 16 local search operators (before filtering, as detailed in Section~\ref{phase1}) to analyze the effect of action space size on the algorithm's performance.
Table~\ref{tab:action_space_comparison} presents the results based on three metrics. The AvS values show that the Q-ALNS with a smaller action space (3-action) generally achieves better solution quality, with a lower median AvS than the 16-action space, indicating more consistent performance. The BS metric, representing the RPD of the best solution found, is also better for the 3-action space configuration, suggesting that a smaller, more focused action space leads to more optimal solutions. Interestingly,  the 16-action space configuration requires less CPU time on average compared to the 3-action space. This may be due to the larger action space allowing for more diverse and potentially faster exploration paths, while the smaller action space, although producing higher quality solutions, may require more local search steps, increasing computation time. 

These findings highlight the trade-off between action space size and algorithm performance. A well-balanced action space is crucial for maintaining a good balance between exploration and exploitation. While a smaller action space with better-performing local search operators can improve solution quality, it may increase computational effort. Conversely, a larger action space can reduce computation time but may lead to less optimal solutions. 

\begin{table}[htbp]
    \centering
    \scriptsize
    \caption{Comparison of Q-ALNS with Different Action Spaces}    \label{tab:action_space_comparison}
    \begin{tabularx}{\textwidth}{c c *{3}{>{\centering\arraybackslash}X} *{3}{>{\centering\arraybackslash}X}}
        \toprule
        \multicolumn{2}{c}{\textbf{Instance}} & \multicolumn{3}{c}{\textbf{Q-ALNS with 16-action space}} & \multicolumn{3}{c}{\textbf{Q-ALNS with 3-action space}} \\
        \cmidrule(r){1-2} \cmidrule(r){3-5} \cmidrule(r){6-8}
        \textbf{Dock} & \textbf{Truck} & \textbf{AvS} & \textbf{BS} & \textbf{T(s)} & \textbf{AvS} & \textbf{BS} & \textbf{T(s)} \\
        \midrule
        3 & 20 & \cellcolor{gray!30}\textbf{0.127} & 0.017 & 17.3 & 0.256 & \cellcolor{gray!30}\textbf{0.008} & 28.7 \\
        4 & 30 & 0.231 & 0.134 & 34.7 & \cellcolor{gray!30}\textbf{0.179} & \cellcolor{gray!30}\textbf{0.080} & 40.4 \\
        5 & 50 & 0.139 & 0.061 & 76.4 & \cellcolor{gray!30}\textbf{0.065} & \cellcolor{gray!30}\textbf{0.023} & 118.9 \\
        6 & 60 & 0.141 & 0.117 & 98.5 & \cellcolor{gray!30}\textbf{0.075} & \cellcolor{gray!30}\textbf{0.029} & 154.1 \\
        7 & 70 & 0.574 & 0.330 & 162.7 & \cellcolor{gray!30}\textbf{0.367} & \cellcolor{gray!30}\textbf{0.106} & 216.4 \\
        8 & 100 & 0.063 & 0.038 & 376.1 & \cellcolor{gray!30}\textbf{0.030} & \cellcolor{gray!30}\textbf{0.012} & 412.2 \\
        8 & 130 & 0.053 & 0.037 & 566.0 & \cellcolor{gray!30}\textbf{0.030} & \cellcolor{gray!30}\textbf{0.016} & 702.2 \\
        9 & 160 & 0.023 & 0.015 & 953.4 & \cellcolor{gray!30}\textbf{0.015} & \cellcolor{gray!30}\textbf{0.003} & 1128.4 \\
        9 & 180 & 0.034 & 0.023 & 1288.3 & \cellcolor{gray!30}\textbf{0.015} & \cellcolor{gray!30}\textbf{0.006} & 1401.4 \\
        10 & 200 & 0.025 & 0.020 & 1645.2 & \cellcolor{gray!30}\textbf{0.014} & \cellcolor{gray!30}\textbf{0.006} & 1973.5 \\
        [1ex] 
        \multicolumn{2}{c}{Average} & 0.141 & 0.079 & 521.9 & \cellcolor{gray!30}\textbf{0.105} & \cellcolor{gray!30}\textbf{0.029} & 617.6 \\
        \bottomrule
    \end{tabularx}
    \label{action_space_comparison}
\end{table}

\subsubsection*{Reward function}

This section analyzes the impact of the reward mechanism on the performance of the Q-ALNS. We compare the original 0-1 reward mechanism with the proposed value-based reward mechanism (detailed in Section \ref{sec4.2}) to understand how different reward structures affect the algorithm's performance.
The results in Table~\ref{tab:reward_mechanism_comparison} reveal that the value-based reward mechanism achieves slightly better solution quality compared to the original 0-1 reward mechanism, with a lower median AvS indicating more consistent performance. The BS metric, representing the best solution found, is also better for the value-based reward mechanism, suggesting that this reward structure leads to more optimal solutions. Additionally, the value-based reward mechanism requires significantly less CPU time on average, likely due to its more efficient guidance of the search process, which reduces the overall computation time needed to find high-quality solutions. 

These findings highlight the effectiveness of the value-based reward mechanism in improving both solution quality and computational efficiency. While the original 0-1 reward mechanism achieves acceptable performance, the value-based reward mechanism offers a better balance between solution quality and computational effort.

\begin{table}[htbp]
    \centering
    \scriptsize
    \caption{Comparison of Q-ALNS with 0-1 Reward Mechanisms}
    \begin{tabularx}{\textwidth}{c c *{3}{>{\centering\arraybackslash}X} *{3}{>{\centering\arraybackslash}X}}
        \toprule
        \multicolumn{2}{c}{\textbf{Instance}} & \multicolumn{3}{c}{\textbf{Original 0-1 reward mechanism}} & \multicolumn{3}{c}{\textbf{ Value-based reward mechanism}} \\
        \cmidrule(r){1-2} \cmidrule(r){3-5} \cmidrule(r){6-8}
        \textbf{Dock} & \textbf{Truck} & \textbf{AvS} & \textbf{BS} & \textbf{T(s)} & \textbf{AvS} & \textbf{BS} & \textbf{T(s)} \\
        \midrule
        3 & 20 & 0.258 & 0.025 & 26.8 & \cellcolor{gray!30}\textbf{0.256} & \cellcolor{gray!30}\textbf{0.008} & 28.7 \\
        4 & 30 & \cellcolor{gray!30}\textbf{0.062} & \cellcolor{gray!30}\textbf{0.062} & 76.6 & 0.179 & 0.080 & 40.4 \\
        5 & 50 & 0.092 & 0.064 & 175.5 & \cellcolor{gray!30}\textbf{0.065} & \cellcolor{gray!30}\textbf{0.023} & 118.9 \\
        6 & 60 & 0.093 & 0.040 & 238.9 & \cellcolor{gray!30}\textbf{0.075} & \cellcolor{gray!30}\textbf{0.029} & 154.1 \\
        7 & 70 & 0.477 & 0.388 & 285.1 & \cellcolor{gray!30}\textbf{0.367} & \cellcolor{gray!30}\textbf{0.106} & 216.4 \\
        8 & 100 & 0.031 & \cellcolor{gray!30}\textbf{0.006} & 743.9 & \cellcolor{gray!30}\textbf{0.030} & 0.012 & 412.2 \\
        8 & 130 & 0.033 & 0.021 & 1172.2 & \cellcolor{gray!30}\textbf{0.030} & \cellcolor{gray!30}\textbf{0.016} & 702.2 \\
        9 & 160 & 0.019 & 0.014 & 1857.7 & \cellcolor{gray!30}\textbf{0.015} & \cellcolor{gray!30}\textbf{0.003} & 1128.4 \\
        9 & 180 & 0.020 & 0.011 & 2478.3 & \cellcolor{gray!30}\textbf{0.015} & \cellcolor{gray!30}\textbf{0.006} & 1401.4 \\
        10 & 200 & 0.016 & 0.009 & 3085.4 & \cellcolor{gray!30}\textbf{0.014} & \cellcolor{gray!30}\textbf{0.006} & 1973.5 \\
        [1ex] 
        \multicolumn{2}{c}{Average} &  0.110 & 0.064 & 1014.0 & \cellcolor{gray!30}\textbf{0.105} & \cellcolor{gray!30}\textbf{0.029} & 617.6 \\
        \bottomrule
    \end{tabularx}    \label{tab:reward_mechanism_comparison}
\end{table}

\subsection{Phase 3: Comparison of dock mode decision with pre-set scenarios}

\begin{figure}[H] 
\centering
\includegraphics[scale=0.3]{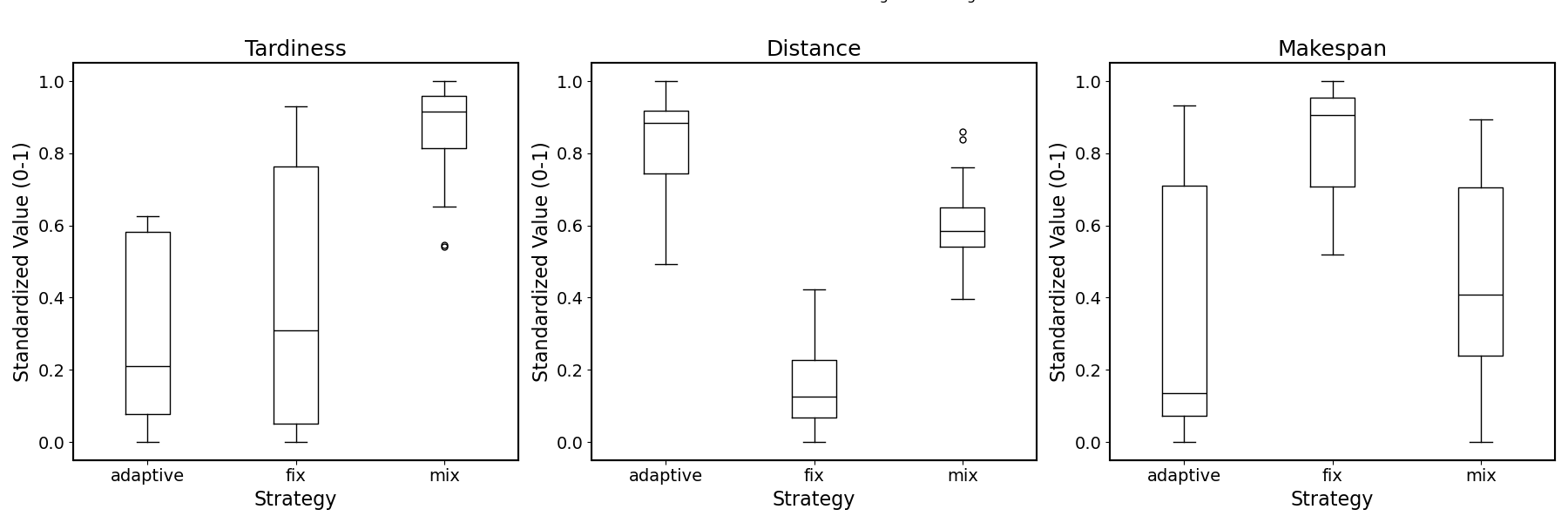} 
\caption{Boxplot of the Adaptive, Fix, and Mix strategies based on tardiness, distance, and makespan} 
\label{fig:strategy} 
\end{figure}

In this phase, we compare the Adaptive, Fix, and Mix dock operation strategies. Figure \ref{fig:strategy} shows boxplots of the three strategies across three objective functions, based on comparisons from 10 instances. To facilitate comparison, the objective function values are normalized using min-max scaling. Our Adaptive strategy performs better in tardiness and makespan. Specifically, the Adaptive strategies achieve an average reduction of 2.2\% and 11.7\% in tardiness and 5.8\% and 0.3\% in makespan compared to the Fix and Mix strategies, respectively. However, this may come at the cost of a relatively higher distance, with increases of 4.2\% and 1.3\%, respectively. 
In the fixed dock mode, the alternating arrangement of loading and unloading docks can create geometric symmetry, thereby minimizing the average distance between storage locations and docks. Conversely, in dock modes with indeterminate or fully mixed usage, scheduling tends to prioritize timely allocation, increasing the complexity of transportation paths. Additionally, the fully mixed mode does not achieve better tardiness, likely due to the truck arrival and departure structure. Our Adaptive strategy demonstrates better adaptability to this structure.

Subsequently, we discuss dock utilization under different strategies. The utilization for a given dock \(i\) is calculated as the ratio of the actual usage time to the total time period. The utilization \(U_i\) of dock \(i\) is determined by \eqnref{eq:U}, where \(n_i\) is the total number of tasks processed on dock \(i\), \(t_{\text{start}, j}\) and \(t_{\text{end}, j}\) are the start and end times of task \(j\), and \(T_{\text{max}}\) is the maximum end time of all tasks across all docks.

\small
\begin{equation}
\label{eq:U}
    U_i = \frac{\sum_{j=1}^{n_i} (t_{\text{end}, j} - t_{\text{start}, j})}{T_{\text{max}}}
\end{equation}

\begin{figure}[ht] 
\centering
\includegraphics[scale=0.5]{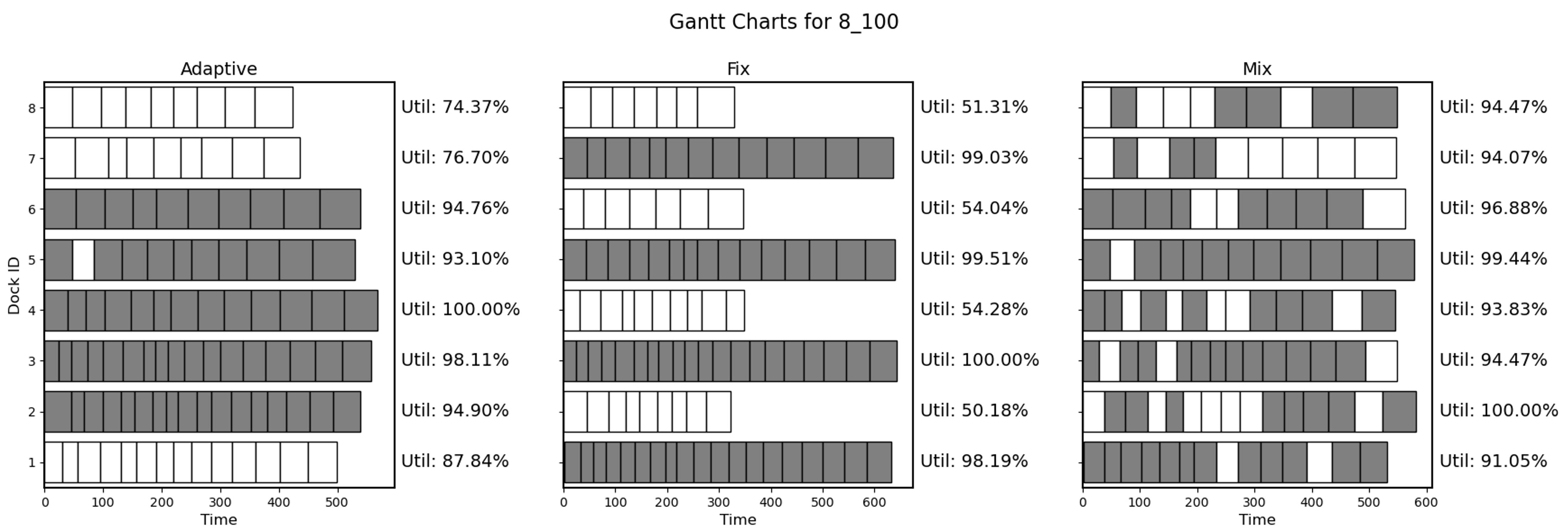} 
\caption{Gantt charts of the Adaptive, Fix, and Mix strategy for the instance 8\_200} 
\label{fig:gant} 
\end{figure}

Figure \ref{fig:gant} shows the Gantt charts for each strategy in a randomly selected instance, where black boxes represent unloading trucks, and white boxes represent loading trucks. The MSM dock significantly improves dock utilization. The Mix strategy consistently maintains high utilization but increases scheduling complexity. The Adaptive strategy, with an average utilization of 89.97\%, outperforms the Fix strategy, which has an average utilization of 75.82\%. Furthermore, the Adaptive strategy achieves this high utilization with only 37.5\% of docks being mixed-use, comparable to the 100\% mixed docks in the Mix strategy. This result indicates that the Adaptive strategy achieves similar high utilization with less operational complexity, aligning with the findings of \citet{rijal_integrated_2019}. 

\section{Conclusion}
\label{sec7}

This paper investigates the integrated truck assignment and scheduling problem with dock mode decision, extending the flexibility of mixed service mode docks compared to the commonly used pre-set dock mode strategies in existing research. We constructed a mathematical model that integrates truck assignment, scheduling, and dock mode decisions, aiming to minimize tardiness, makespan, and AGV handling distance. To solve this complex combinatorial optimization problem, we proposed a Q-learning-based Adaptive Large Neighborhood Search (Q-ALNS) algorithm in which Q-learning is embedded into ALNS to guide the adaptive operator selection mechanism.

To validate the proposed approach, we conducted a three-phase experiment. These experiments indicate that the performance of the ALNS relies heavily on filtering better-performing operator combinations, which can significantly enhance the algorithm's adaptability to the problem. The Q-ALNS, which incorporates Q-learning, consistently outperforms the original ALNS and other benchmarks in terms of optimality gap and Pareto front quality, with comparable overhead complexity. Especially in large-scale problems, it takes a shorter time to reach the best solution. This validates the competitiveness of the Q-ALNS in balancing exploration and exploitation. The sensitivity analysis revealed that a well-balanced action space and an effective reward mechanism are crucial for optimizing the performance of the Q-ALNS. Additionally, incorporating dock mode decision strategies results in relatively lower objective function values, more balanced and efficient Gantt charts, and higher dock utilization rates, significantly enhancing the operational flexibility and efficiency of the inbound and outbound processes.

Despite the significant benefits achieved, the current model assumes a level of certainty in operational conditions, which might not fully capture the complexities of real-world scenarios. A promising direction for future research is to extend the model to account for uncertainty, enhancing its applicability to more dynamic and complex environments. On the algorithmic side, the integration of Q-learning into the ALNS algorithm has proven effective, and future work could further enhance the intelligence of parameter selection, improving the algorithm’s adaptability and performance. 

\section*{Acknowledgements}

Funding: This work was supported by the National Key Research and Development Program of China [grant number 2021YFB1407003]; and the China Scholarship Council [grant number 202307090057].

\bibliography{reference}
\clearpage
\appendix

\section{Detailed Descriptions and Sources of 
 Local Search Operators}
\label{app2}

The local search operators in ALNS are formed by a combination of destroy, repair, and comprehensive operators. Some operators were adapted from the design ideas in \citet{bodnar_scheduling_2017, rijal_integrated_2019}, but were modified to suit the specifics of our problem. 

Our design of destroy operators follows the most commonly used random and greedy operators found in the ALNS literature. The random destruction operator randomly selects a subset of elements from the current solution to remove, creating diversity in the search process. The greedy destruction operator, on the other hand, targets the removal of elements that are likely to lead to immediate improvements in the solution quality when reinserted.
The destroy operators in the proposed algorithm are as follows:

\begin{itemize}
    \item \textbf{rRd}: Randomly selecting and removing a truck from all available trucks.
    \item \textbf{rMxTar}: Targeting and removing the truck that causes the most significant delay in the current solution.
    \item \textbf{rMxM}: Targeting and removing the truck with the greatest weighted cargo handling distance from all trucks. The weighted cargo handling distance for each truck is calculated using the formula given in Equation (\ref{eq:weighted_distance}).
\end{itemize}

\begin{equation}
    \mu_i = \sum_{k \in D} \sum_{s \in S} \sum_{g \in G} x_{ik} h_{is} u_{sg} m_{kg} \label{eq:weighted_distance}
\end{equation}

The repair strategies considered in this paper include forward, backward, swapping, and inserting. The repair operators are as follows:

\begin{itemize}
    \item \textbf{iBck}: Swap the removed truck with its preceding truck at the same dock.
    \item \textbf{iFwd}: Swap the removed truck with its succeeding truck at the same dock.
    \item \textbf{iSwap}: Depending on whether it is for unloading or loading, swap the removed truck with another randomly selected unloading or loading truck, not limited to the same dock.
    \item \textbf{iUp}: Reallocate the removed truck to a dock with a higher rank based on handling distance.
    \item \textbf{iDown}: Reallocate the removed truck to a dock with a lower rank based on handling distance.
    \item \textbf{iDockInsert}: Attempt to insert the removed truck at all positions in the truck sequence of the same dock, replacing the current solution if improved.
    \item \textbf{iBtwInsert}: Insert the removed truck randomly in the truck sequence of different docks, replacing the current solution if improved.
\end{itemize}

The comprehensive operators, distinct from the destroy and repair operators, independently modify an existing solution. The designed comprehensive operators in this paper include the following three types:

\begin{itemize}
    \item \textbf{riInD2D}: Randomly selects two docks used for unloading ($D_I$) and swaps the trucks they service, while maintaining the sequence of trucks.
    \item \textbf{riOuD2D}: Randomly selects two docks used for loading ($D_O$) and swaps their serviced trucks, keeping the truck order unchanged.
    \item \textbf{riFlxD2D}: In mixed-mode docks ($D_F$), randomly selects two docks and swaps their serviced trucks, again preserving the truck sequence.
\end{itemize}

\section{Sensitive of the Q-ALNS to its parameters}
\label{app0}

\begin{figure}[H] 
\centering
\includegraphics[scale=0.28]{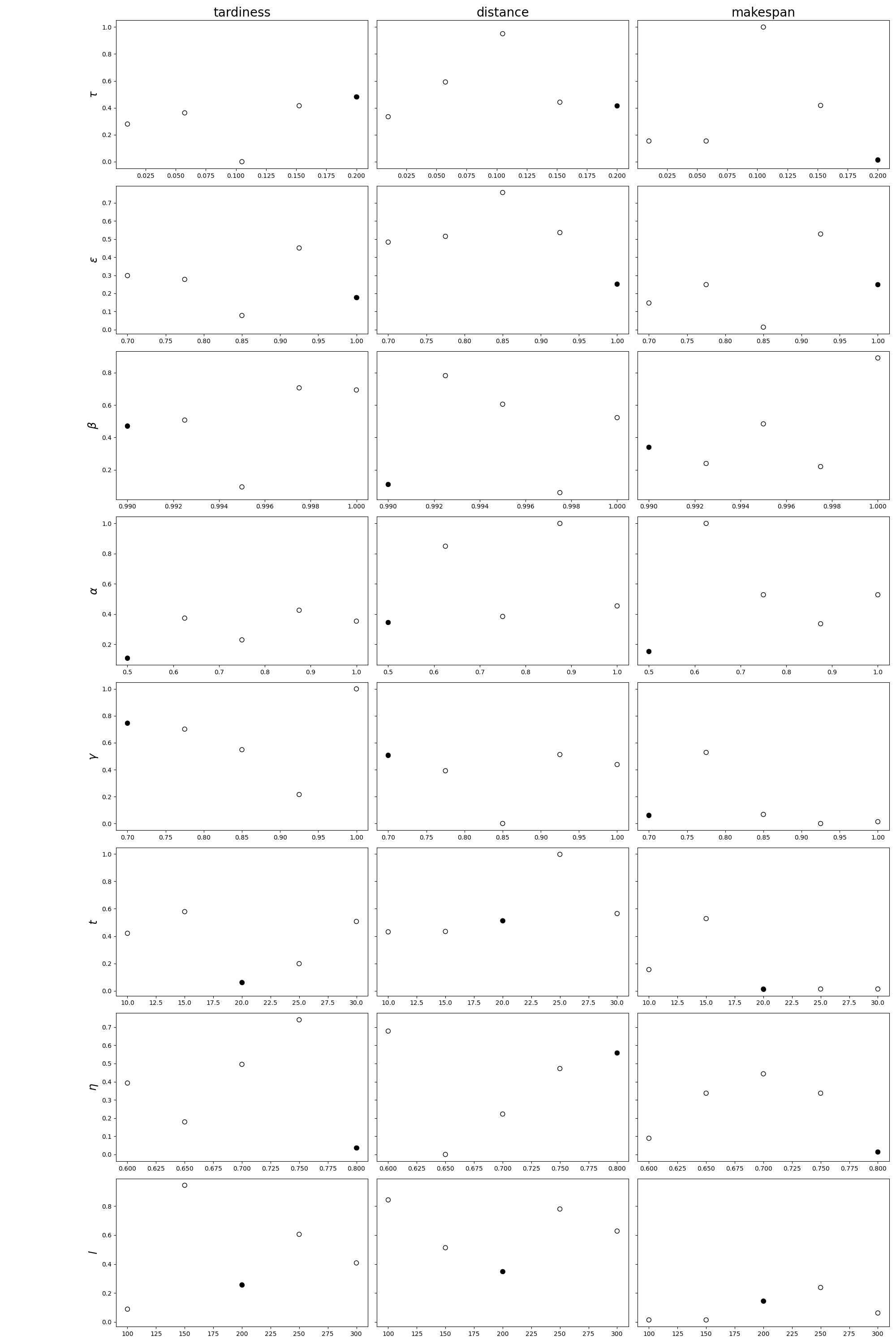} 
\caption{Sensitive of the Q-ALNS to its parameters based on tardiness, distance, and makespan} 
\label{fig:sensitive} 
\end{figure}

\end{document}